\def\paperID{4903}
\def\confName{CVPR}
\def\confYear{2024}

\def\authorBlock{
    Hongwei Zheng\textsuperscript{1,2}, Linyuan Zhou\textsuperscript{1}, Han Li\textsuperscript{2}, 
    Jinming Su\textsuperscript{1}, Xiaoming Wei\textsuperscript{1}, Xiaoming Xu\textsuperscript{1}
    $^{\dagger}$
    \\
    \textsuperscript{1} Meituan \qquad
\textsuperscript{2} Shanghai Jiao Tong University
    \\
    {\tt\small \{zhenghongwei04, zhoulinyuan, sujinming, weixiaoming, xuxiaoming04\}@meituan.com } \\
    {\tt\small\{qingshi9974\}@sjtu.edu.cn }
}

\newif\ifreview 
\newif\ifarxiv \newcommand{\arxiv}{\arxivtrue}
\newif\ifcamera 
\newif\ifrebuttal 

\arxiv 
\pdfoutput=1
\documentclass[10pt,twocolumn,letterpaper]{article}
\ifreview \usepackage[review]{cvpr} \fi
\ifarxiv \usepackage[pagenumbers]{cvpr} \fi
\ifrebuttal \usepackage[rebuttal]{cvpr} \fi
\ifcamera \usepackage{cvpr} \fi

\usepackage{graphicx}	
\usepackage{amsmath}	
\usepackage{amssymb}	
\usepackage{booktabs}
\usepackage{times}
\usepackage{microtype}
\usepackage{epsfig}
\usepackage[table,xcdraw,dvipsnames]{xcolor}
\usepackage{caption}
\usepackage{float}
\usepackage{placeins}
\usepackage{color, colortbl}
\usepackage{stfloats}
\usepackage{enumitem}
\usepackage{tabularx}
\usepackage{xstring}
\usepackage{multirow}
\usepackage{xspace}
\usepackage{url}
\usepackage{subcaption}
\usepackage{xcolor}
\usepackage[hang,flushmargin]{footmisc}

\usepackage{algorithm}
\usepackage{algorithmic}
\usepackage{afterpage}
\usepackage{bm}
\usepackage{setspace}

\ifcamera \usepackage[accsupp]{axessibility} \fi

% ...

\ifarxiv  \fi

\newcommand{\R}[1]{{%
    \textbf{%
        \ifstrequal{#1}{1}{\textcolor{red}{R#1}}{%
        \ifstrequal{#1}{2}{\textcolor{blue}{R#1}}{%
        \ifstrequal{#1}{3}{\textcolor{magenta}{R#1}}{%
        \ifstrequal{#1}{4}{\textcolor{teal}{R#1}}{%
                           \textcolor{cyan}{R#1}%
        }}}}%
    }%
}}

\usepackage{xr-hyper}

\makeatletter
\newcommand*{\addFileDependency}[1]{
  \typeout{(#1)}
  \@addtofilelist{#1}
  \IfFileExists{#1}{}{\typeout{No file #1.}}
}

\makeatother
\newcommand*{\myexternaldocument}[1]{
    \externaldocument{#1}
    \addFileDependency{#1.tex}
    \addFileDependency{#1.aux}
}

\definecolor{cvprblue}{rgb}{0.21,0.49,0.74}
\usepackage[pagebackref,breaklinks,colorlinks,citecolor=cvprblue]{hyperref}
\usepackage[capitalize]{cleveref}
\crefname{section}{Sec.}{Secs.}
\crefname{table}{Table}{Tables}
\crefname{figure}{Fig.}{Figs.}

\myexternaldocument{_supplementary}
\begin{document}
\title{BEM: Balanced and Entropy-based Mix \\
for Long-Tailed Semi-Supervised Learning}
\author{\authorBlock}

\twocolumn[{%
\maketitle
\vspace{-20pt}
\begin{figure}[H]
\hsize=\textwidth 
\centering
\subfloat[Consistent class distribution]{\includegraphics[scale=0.194]{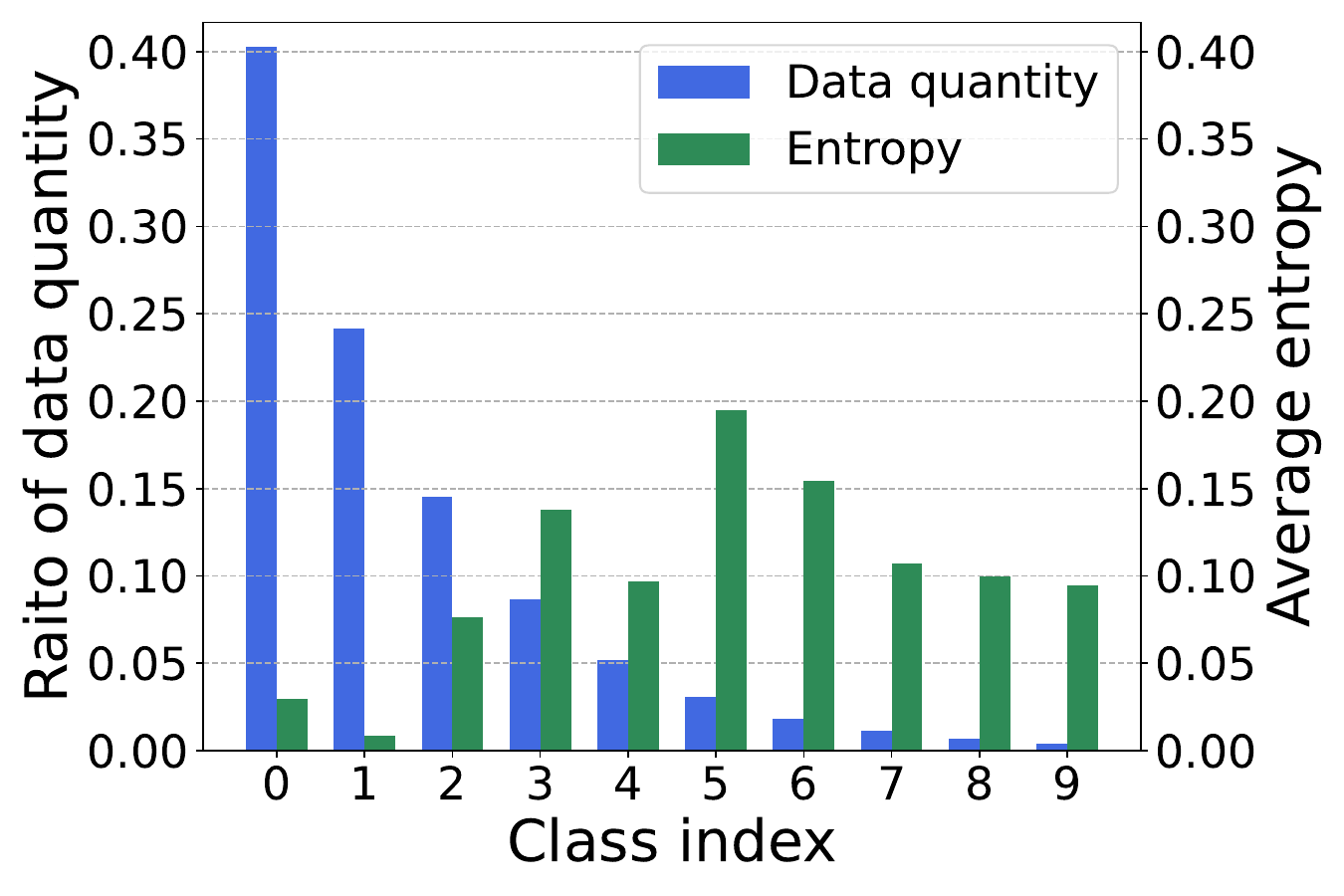}}
\subfloat[Uniform class distribution]{\includegraphics[scale=0.194]{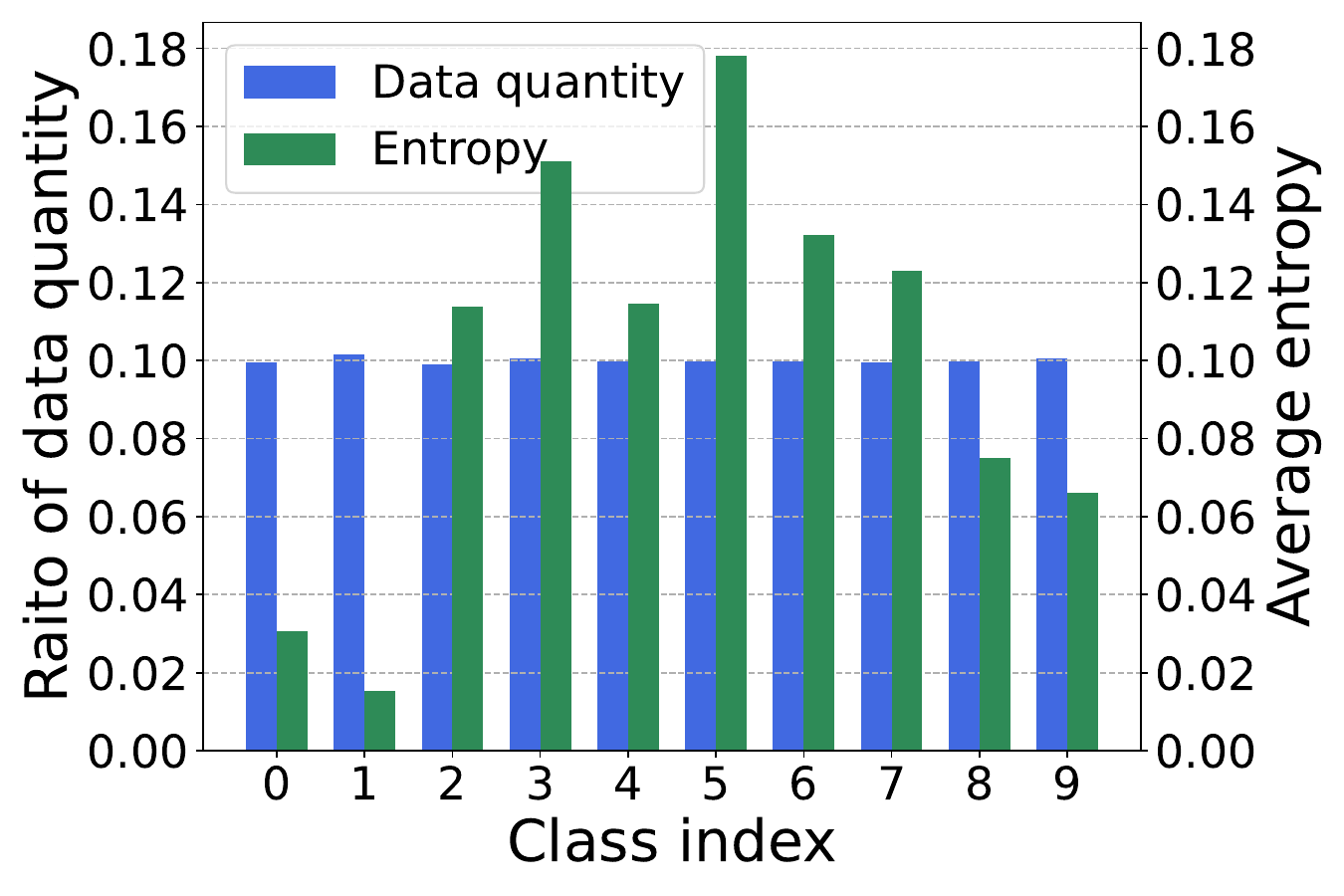}}
\subfloat[Reversed class distribution]{\includegraphics[scale=0.194]{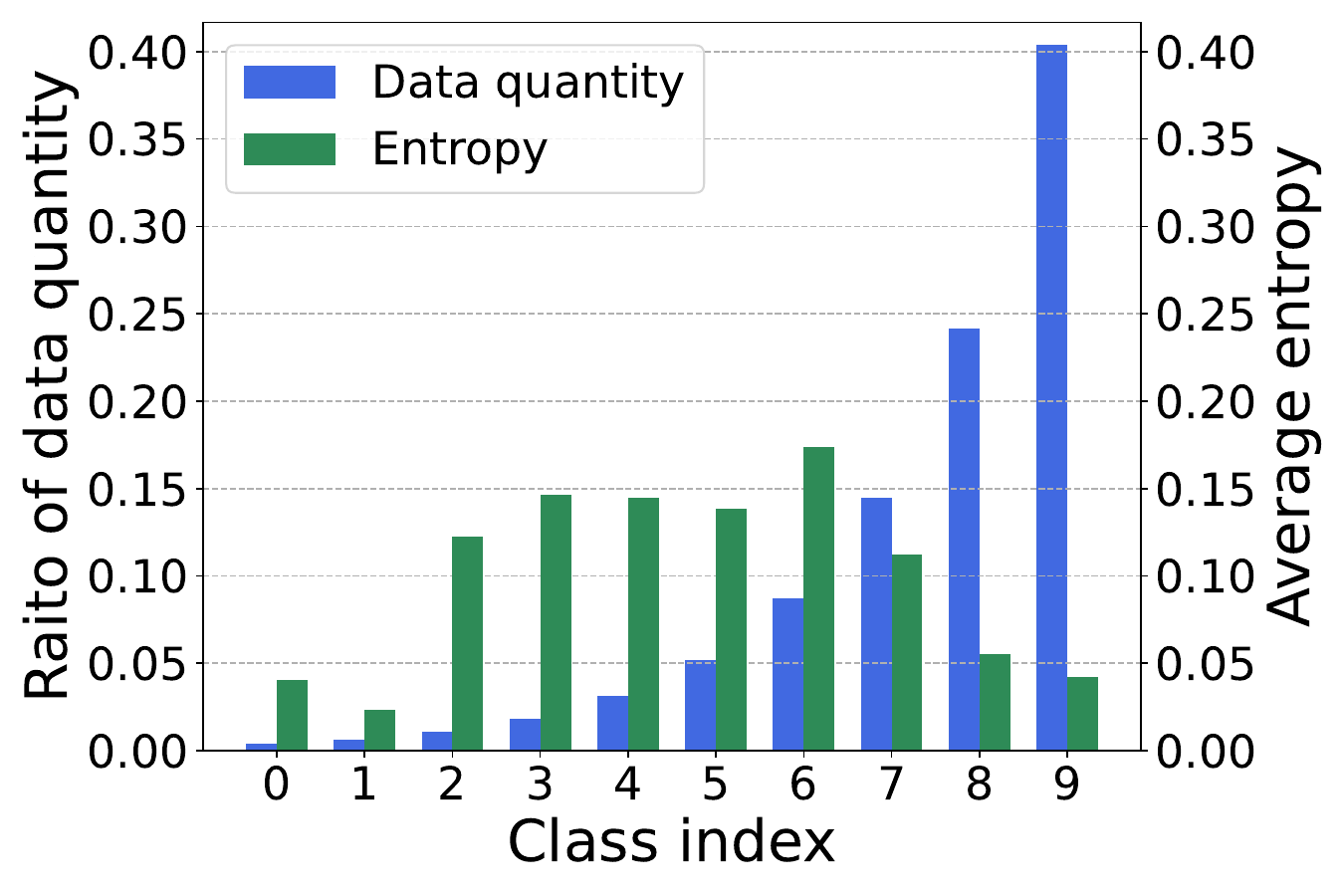}}
\subfloat[Test accuracy gain]{\includegraphics[scale=0.194]{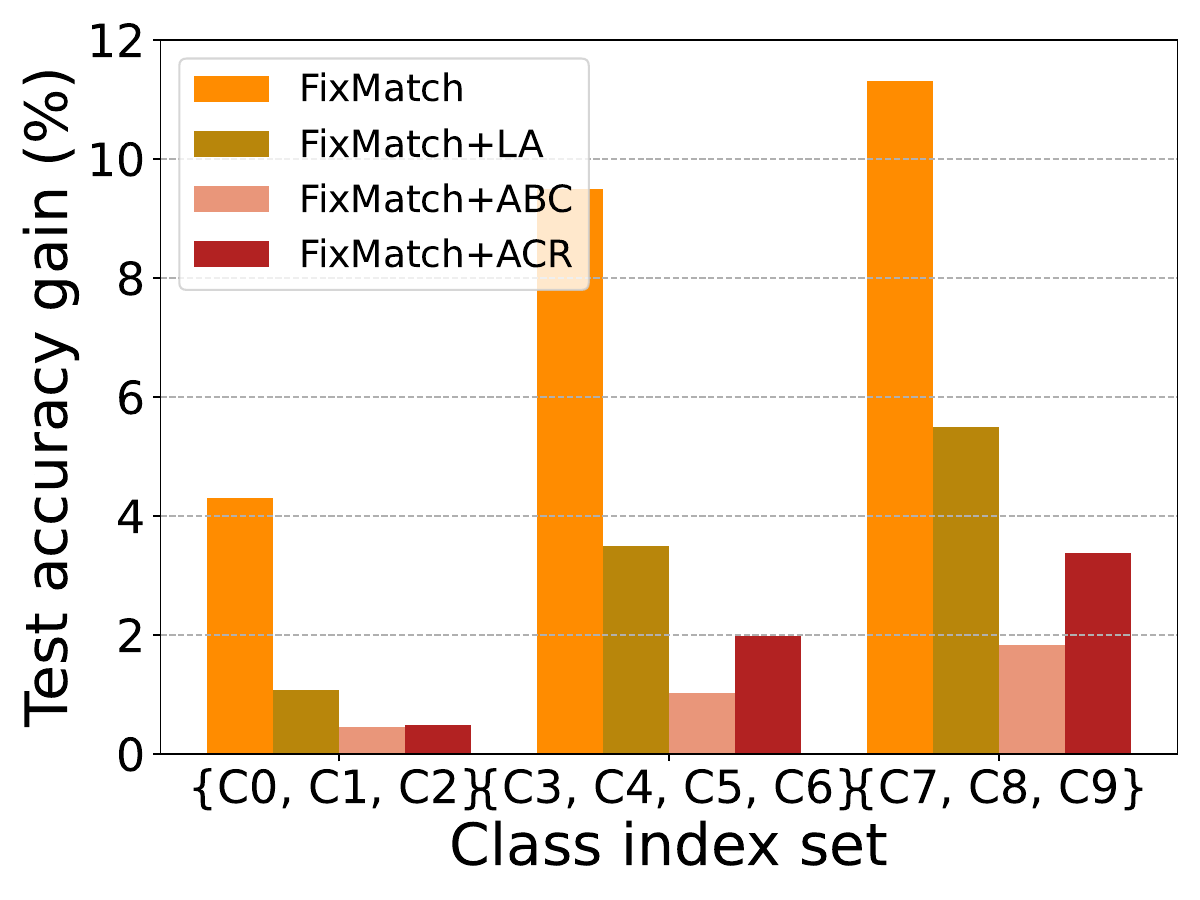}}\\
\caption{Experimental results on CIFAR10-LT~\cite{krizhevsky2009learning}. \textbf{(a)-(c):} Class distribution of unlabeled data quantity and entropy for three typical settings,
which have the same labeled data quantity distribution but differ in unlabeled ones.
Both the data quantity and entropy are the statistical averages within one epoch after model convergence.
Unexpected discrepancies are observed across all settings between the distribution of data quantity and entropy, particularly for head and tail classes.  Notably, classes 3-6 exhibit the highest entropy, indicating greater uncertainty.
\textbf{(d):} Test accuracy gain brought by BEM for various LTSSL frameworks in consistent setting.}
\label{fig:motivation}
\end{figure}
}]

\begin{abstract}
Data mixing methods play a crucial role in semi-supervised learning (SSL), but their application is unexplored in long-tailed semi-supervised learning (LTSSL). The primary reason is that the in-batch mixing manner fails to address class imbalance. Furthermore, existing LTSSL methods mainly focus on re-balancing data quantity but ignore class-wise uncertainty, which is also vital for class balance. For instance, some classes with sufficient samples might still exhibit high uncertainty due to indistinguishable features. To this end, this paper introduces the Balanced and Entropy-based Mix (BEM), a pioneering mixing approach to re-balance the class distribution of both data quantity and uncertainty. Specifically, we first propose a class balanced mix bank to store data of each class for mixing. This bank samples data based on the estimated quantity distribution, thus re-balancing data quantity. Then, we present an entropy-based learning approach to re-balance class-wise uncertainty, including entropy-based sampling strategy, entropy-based selection module, and entropy-based class balanced loss. Our BEM first leverages data mixing for improving LTSSL, and it can also serve as a complement to the existing re-balancing methods. Experimental results show that BEM significantly enhances various LTSSL frameworks and achieves state-of-the-art performances across multiple benchmarks. 
\end{abstract}
\let \thefootnote \relax \footnotetext{ $\dagger$ Corresponding Author}

\section{Introduction}




Semi-supervised learning (SSL) capitalizes on unlabeled data to reduce the cost of data labeling and boost the performance of models~\cite{van2020survey, ouali2020overview, shim2020data,grandvalet2004semi,laine2016temporal}. 
The general paradigm of most approaches is to randomly generate two views of an image with various augmentation methods and then use the output of one as the pseudo label to supervise the other~\cite{ sohn2020fixmatch, zhang2021flexmatch,chen2023softmatch}. 
As a simple and effective augmentation technique introduced in supervised learning~\cite{he2019bag,huang2017snapshot,berthelot2018understanding,verma2019manifold}, data mixing is widely used in SSL algorithms~\cite{berthelot2019mixmatch, berthelot2019remixmatch, verma2022interpolation}, further enhancing model generalization and performance.

 

However, most existing SSL algorithms assume a balanced dataset, ignoring the real-world prevalence of long-tailed class distributions~\cite{bengio2015sharing, cui2019class, zhang2023deep, zhang2021distribution, hong2022long,hong2023harmonizing}.
To deal with this class imbalance scenario,
various long-tailed semi-supervised learning (LTSSL) methods have been proposed,
such as re-sampling~\cite{wei2021crest}, logit alignment~\cite{menon2020long, wei2023towards}, and pseudo label alignment \cite{oh2022daso, lee2021abc}.
Nevertheless, data mixing is rarely explored in LTSSL.
The primary reason is that existing data mixing methods (\emph{e.g.} MixUp~\cite{zhang2017mixup}, CutMix~\cite{yun2019cutmix}, and SaliencyMix~\cite{uddin2020saliencymix}) often perform random mixing within a batch. Consequently, the infrequent tail classes may not be adequately sampled when the batch size is small. This hinders  a balanced class distribution,
which is crucial for LTSSL.


To make data mixing suitable for LTSSL, 
let us consider two key questions: \textbf{i) How can we apply data mixing to effectively re-balance the data quantity for each class?
ii) Is it sufficient to solely focus on re-balancing data quantity to achieve class balance?
}  
For the second question,
we notice that previous LTSSL methods mainly focus on addressing the issue of long-tailed class distribution in terms of data quantity. These methods ignore the fact that class performance also depends on class-wise uncertainty~\cite{li2022trustworthy, li2022nested, samuel2021distributional, li2023pose}, which is associated with the training difficulty for each class.
For instance, some classes with sufficient samples may still encounter high training difficulty due to the high uncertainty induced by indistinguishable features.
As shown in Fig.~\ref{fig:motivation} (a)-(c),  we quantify the uncertainty by entropy~\cite{li2022nested} and compare this class-wise entropy with data quantity under three typical settings~\cite{wei2023towards}.
The results reveal a \textbf{significant disparity} between the class distribution of entropy and data quantity across all settings. This finding emphasizes the limitation of solely re-balancing data quantity, as it does not consider classes with high uncertainty, ultimately limiting performance improvement.
Thus, it is crucial to also address the re-balancing of class-wise uncertainty, \emph{i.e.} entropy.

To tackle the above problems, this paper presents a novel data mixing paradigm, called Balanced and Entropy-based Mix (BEM), for LTSSL. Specifically, 
we first introduce a simple mixing strategy, named as CamMix, which has a strong localization capability to avoid redundant areas for mixing.
Then, we establish a class balanced mix bank (CBMB) to store and sample class-wise data for mixing. The sampling function follows the estimated class distribution of data quantity and we adopt the effective number~\cite{cui2019class} to represent the realistic data quantity of each class.
Our CamMix incorporated CBMB can effectively re-balance the class-wise data quantity in an end-to-end optimized manner, which can not be achieved by the re-sampling methods~\cite{kang2019decoupling,cao2019learning,zhou2020bbn,wei2021crest} with the complex training procedures.

Further, we present a novel entropy-based learning approach to re-balance class-wise uncertainty. 
Entropy-based sampling strategy (ESS) integrates class-wise entropy into the quantity-based sampling function.
In addition, entropy-based selection module (ESM) adaptively
determines the sampled data ratio between labeled and unlabeled data during mixing
to manage the trade-off between guiding high-uncertainty unlabeled data~\cite{arazo2020pseudo, wang2022freematch} with confident labeled data and maximizing the utilization of unlabeled data.
Finally, we incorporate the class balanced loss~\cite{cui2019class} with class-wise entropy to form entropy-based class balanced (ECB) loss. 

We highlight that our BEM is the first method that leverages data mixing to enhance LTSSL. Our results demonstrate that BEM can effectively complement existing re-balancing methods by boosting their performance across several benchmarks.
As shown in Fig.~\ref{fig:motivation} (d), our method enhances FixMatch~\cite{sohn2020fixmatch}, FixMatch+LA~\cite{menon2020long}, FixMatch+ABC~\cite{lee2021abc}, FixMatch+ACR~\cite{wei2023towards}, achieving to 11.8\%, 4.4\%, 1.4\% and 2.5\% average gains on test accuracy, respectively.
Additionally, BEM proves to be a versatile framework, performing well across different data distributions, diverse datasets, and various SSL learners.

\section{Related Work}

\noindent\textbf{Data mixing.} 
MixUp~\cite{zhang2017mixup} and CutMix~\cite{yun2019cutmix} are typical data mixing methods used in various computer vision tasks.
While performing mixing at element-wise and region-wise levels respectively, they share a common limitation of neglecting class content, thus introducing substantial redundant context irrelevant to class content~\cite{du2023global, park2022majority}.
To achieve class balance in LTSSL, it is essential to ensure that the selected region for mixing contains related class content and avoids redundancy.
SaliencyMix~\cite{uddin2020saliencymix} alleviates this issue by using a saliency map to ensure  that selected regions contain class content, but the resulting region is still too coarse to avoid numerous redundant areas.
In our paper, CamMix achieves tighter localization of class regions to minimize redundant areas, which is particularly well-suited for LTSSL.

\noindent\textbf{Data mixing in semi-supervised learning.} 
Data mixing is crucial in SSL,
enhancing model performance by creating diverse training samples.
For instance, MixMatch~\cite{berthelot2019mixmatch} utilizes MixUp~\cite{zhang2017mixup} as the data mixing technique to learn a robust model. ReMixMatch~\cite{berthelot2019remixmatch} adds distribution alignment and augmentation anchoring to the MixMatch framework.
ICT~\cite{verma2022interpolation} employs the mean teacher model and implements MixUp on unsupervised samples.
Despite widely used in SSL algorithms, almost no methods in LTSSL apply data mixing.
This is mainly due to the limitation of employing in-batch mixing, which fails to address the class imbalance problem.
Our method stands out as the first to incorporate data mixing in LTSSL.

\noindent\textbf{Long-tailed semi-supervised learning.}
LTSSL is gaining attention due to its real-world applicability. For example, CReST~\cite{wei2021crest} refines the model by iteratively enriching the labeled set with high-quality pseudo labels in multiple rounds. ABC~\cite{lee2021abc} uses an auxiliary balanced classifier, trained by down-sampling majority classes. DASO~\cite{oh2022daso} mitigates class bias by adaptively blending linear and semantic pseudo labels. ACR~\cite{wei2023towards}, the current state-of-the-art method, proposes a dual-branch network and dynamic logit adjustment.
However, none of these methods utilizes data mixing to further enhance their performance as in SSL.
CoSSL~\cite{fan2022cossl} uses MixUp at the feature level for minority classes and decouples representation learning and classifier learning. However, this feature mixing method assumes identical class distributions for labeled and unlabeled data and requires a complex training approach. Our study considers non-ideal class distributions and designs a simple image-level mixing method in an end-to-end training framework.


\section{Preliminaries}

\begin{figure*}[!htp]
    \centering
    \includegraphics[width=0.9\linewidth]{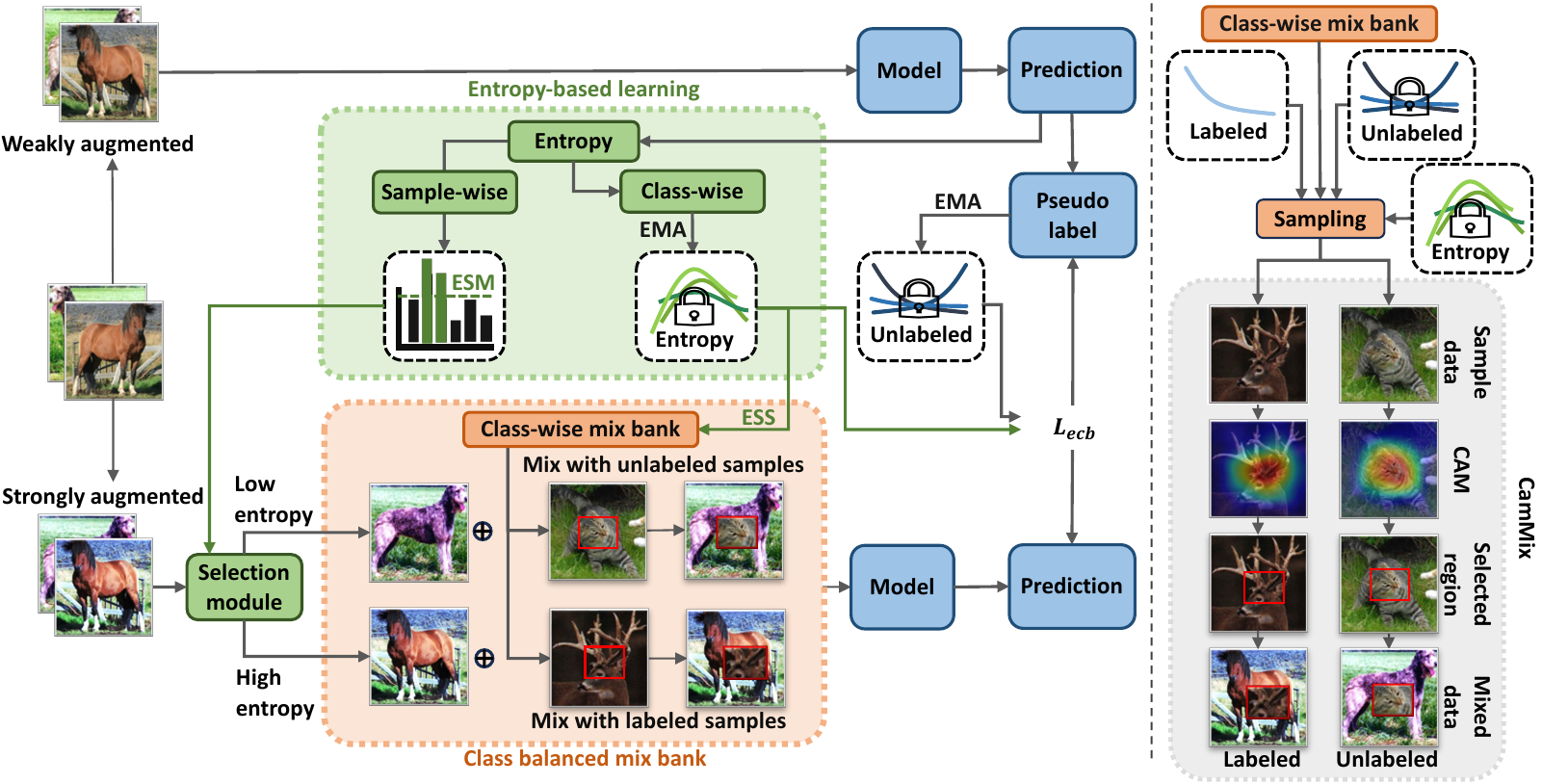}
    \caption{\textbf{Left}: 
    The overview of Balanced and Entropy-based Mixing (BEM), incorporating with FixMatch~\cite{sohn2020fixmatch} as an example in this figure.  BEM consists of two sub-modules: class balanced mix bank (CBMB) and entropy-based learning (EL).
    CBMB re-balances data quantity through the proposed CamMix, guided by a class-balanced sampling function.
    EL further re-balances class-wise uncertainty using three techniques: entropy-based sampling strategy (ESS), entropy-based selection module (ESM) and entropy-based class balanced loss ($L_{ecb}$). 
    \textbf{Right}: The sampling and CamMix process of BEM. 
    The sampling process considers both the class distribution of data quantity and uncertainty, which are estimated on the fly.
    CamMix extracts the bounding box from the high response area of the CAM to form mixed data.
    (The lock icon denotes the unknown distribution that needs estimation, and the $\oplus$ icon denotes the process of CamMix.).
   }
    \label{fig:framework}
    \vspace{-10pt}
\end{figure*}

\noindent\textbf{Semi-supervised learning.} In SSL, the training data consists of labeled data $X=\{(x_n, y_n)\}_{n=1}^N$ and unlabeled data $U=\{u_m\}_{m=1}^M$. Here, $x_n$ and $u_m$ are training samples, $y_n$ is the ground truth, $N$ and $M$ denote the quantity of labeled data and unlabeled data, respectively. 
A representative framework of SSL is FixMatch~\cite{sohn2020fixmatch}, which utilizes unlabeled data with the \emph{Weak and Strong Augmentation}. For an unlabeled sample $u_m$, it first takes a \emph{weakly}-augmented version of $u_m$ as 
the input of the model $f(\cdot)$ to compute the prediction. Then, it uses $q_m =\mathrm{argmax}(f(A_w(u_m)))$ as one-hot pseudo label, while applying the prediction from a \emph{strongly}-augmented of $u_m$ to calculate the cross entropy loss $L_u$:
\begin{equation}
    L_u = \sum_{m=1}^{B}\mathbb{I}(\max(f(A_w(u_m)))>\tau)\underbrace{\mathcal{H}(f(A_s(u_m)), q_m)}_{L_{cls}},
\end{equation}
  where $B$ denotes the batch size, $\mathcal{H}(\cdot)$ is cross entropy and $L_{cls}$ denotes original classification loss term.
  $\mathbb{I}(\max(q_m)>\tau)$ is the 
 mask to filter low-confidence pseudo label with a threshold of $\tau$, abbreviated as $M_u(\cdot)$ in the following part. $A_w$ and $A_s$ denotes the \emph{weak augmentation} (\emph{e.g.}, random crop and flip) and \emph{strong augmentation} (\emph{e.g.}, RandAugment~\cite{cubuk2020randaugment} and Cutout~\cite{devries2017improved}), respectively.
  

\noindent\textbf{Long-tailed semi-supervised learning.} 
In LTSSL, a dataset with a long-tailed distribution is characterized by the minority of classes possessing a large number of samples, while the majority of classes contain only a few samples.
Given $C$ classes across the dataset, $N_c$ represents the quantity of labeled data for class $c$. Without loss of generality, we assume that $N_1 \geq N_2 \geq \cdot\cdot\cdot \geq N_C$ and the imbalanced ratio is denoted by $\gamma_l=N_1/N_C$. Similarly, we can denote the quantity of unlabeled data as $M_c$ for class $c$ and the imbalanced ratio as $\gamma_u=\max_c M_c/\min_c M_c$. 

\section{Balanced and Entropy-based Mix (BEM)}

The Balanced and Entropy-based Mix (BEM) is a plug-and-play method based on the existing SSL framework. Fig.~\ref{fig:framework} shows the overview of BEM, incorporating FixMatch~\cite{sohn2020fixmatch} as an example. Specifically, our entropy-based learning (EL) takes the prediction of the weakly augmented samples as input to perform entropy-based sampling and selection. Then, strongly augmented samples are mixed with data from class balanced mix bank (CBMB) using our CamMix. Based on the estimated distribution of data quantity and uncertainty, we employ the entropy-based class balance (ECB) loss $L_{ecb}$ to train the overall framework.
Please refer to Appendix~\textcolor{red}{C} for the pseudo-code of BEM.
\subsection{CamMix}

Most data mixing methods, such as MixUp~\cite{zhang2017mixup} and CutMix~\cite{yun2019cutmix}, lack the localization ability for class re-balancing.
Although SaliencyMix~\cite{uddin2020saliencymix} has initial localization ability, it still tends to extract excessive redundant context.
To this end, we propose CamMix to replace the saliency map of SaliencyMix with Class Activation Map (CAM)~\cite{zhou2016learning} to achieve more accurate localization. Specifically, we feed images into the prediction model (\emph{i.e.} ResNet50~\cite{he2016deep}) to generate the CAM, where the last layer of the third block of ResNet50 is used as the CAM layer. The resulting CAM is used to extract the largest connected region using a threshold of $\tau_c$. Finally, we obtain the bounding box of this region and paste the corresponding patch onto the original image. The pseudo-code of CamMix can be found in the Appendix~\textcolor{red}{C}.

\subsection{Class Balanced Mix Bank (CBMB)}
Previous in-batch data mixing methods used in SSL are limited to increasing the data quantity of tail classes, thus failing to re-balance the class distribution.
To address this issue, we further propose a class balanced mix bank (CBMB) that stores samples for each class and adequately selects samples to be mixed based on a prior-based class-balancing rule. In essence, the more frequent a class, the more samples are used in the data mixing process. As noted in~\cite{cui2019class}, there is overlap in the data, necessitating the use of the effective number $E_c$ to measure the realistic class distribution of data quantity:
\begin{equation}
    E_c = \frac{1-\beta^{N_c}}{1-\beta},
    \label{equ:2}
\end{equation}
where $N_c$ represents the data quantity of class $c$, while the hyper-parameter $\beta$ is set to 0.999 in our experiments.

The effective number of labeled data, denoted as $E_c^x$, can be obtained directly using Eq.~\ref{equ:2}.  
As the class distribution of unlabeled data is unknown, we estimate it using a simple yet effective approach. Specifically, at each iteration $t$, we obtain the class distribution of the pseudo label $d_c^{ut}$ and update the class distribution of the entire unlabeled dataset $d_c^u$, using an Exponential Moving Average (EMA) approach once the training status stabilizes.
\begin{equation}
    d_c^u \leftarrow \lambda_d d_c^u + (1-\lambda_d) d_c^{ut},
    \label{equ:3}
\end{equation}
where $\lambda_d$ denotes the EMA weight. To obtain the effective number of unlabeled data for each class $E_c^u$, we substitute the class-wise data quantity $N_c^u = M d_c^u$ into Eq.~\ref{equ:2}, where $M$ is the quantity of entire unlabeled dataset.  Then, we obtain the effective number of total data for each class by $E_c = E_c^x + E_c^u$ and perform our CamMix using the initial sampling function as follows:
\begin{equation}
  s_c = \frac{F_c}{\sum_{c=1}^C F_c},
  \label{equ:4}
\end{equation}
where $F_c =1 / E_c $ and  $s_c$ denotes the sampling probability for class $c$.
By accurately estimating the class distribution of the dataset, we can enhance the precision of mixed data sampling. 
Our data mixing achieves class balance among training samples, equivalent to re-sampling during training.
\subsection{Entropy-based Learning (EL)}
In the previous section, we re-balance training samples to initially alleviate the long-tail distribution problem. However, class balance does not only depend on data quantity.
Class-wise uncertainty, which can be quantified by entropy, is also vital for class performance as it reflects training difficulty.
Thus, we propose an entropy-based learning approach to re-balance class-wise entropy, including entropy-based sampling strategy (ESS), 
entropy-based selection module (ESM) and entropy-based class balanced (ECB) loss.  

\noindent\textbf{Entropy-based Sampling Strategy.}
To consider class-wise uncertainty in the sampling process, we define the class-wise entropy $e_{c}^x$ and $e_{c}^u$ for the entire labeled and unlabeled dataset, and 
update them in EMA manner by using the average entropy $e_c^{xt}$ and $e_c^{ut}$ at each training iteration $t$ as follows:
\vspace{-10pt}
\begin{equation}
\begin{aligned}
    e_{c}^{xt} &= \frac{1}{N_c^t} \sum_{n=1}^{N_c^t} \sum_{c=1}^{C} -f_c(A_w(x_n)) \log(f_c(A_w(x_n)))  \\
    e_{c}^{ut} &= \frac{1}{M_c^t} \sum_{m=1}^{M_c^t } \sum_{c=1}^{C} -f_c(A_w(u_m)) \log(f_c(A_w(u_m))),
  \label{equ:5}
\end{aligned}
\end{equation}
\begin{equation}
\begin{aligned}
e_{c}^{x} &\leftarrow \lambda_e e_{c}^{x} + (1-\lambda_e) e_{c}^{xt}  \\
    e_{c}^{u} &\leftarrow \lambda_e e_{c}^{u} + (1-\lambda_e) e_{c}^{ut},
  \label{equ:6}
\end{aligned}
\end{equation}
where $N_c^t$ and $M_c^t$ represent the data quantity within one batch belonging to class $c$ according to the ground truth and pseudo label respectively, and $\lambda_e$ denotes the EMA weight. It's worth noting that we start estimating the entropy of data once the training status stabilizes.
Then, we obtain the total class-wise entropy, \emph{i.e.} $e_c =e_{c}^u+ e_{c}^x$, and subsequently compute the final sampling probability $\hat{s}_c$:
\begin{equation}
    \hat{s}_c = \delta(\alpha s_c + (1-\alpha)s_c^\prime),
\label{equ:alpha}
\end{equation}
where $s_c^\prime$ is the normalization of $e_c$, denoted as $s_c^\prime =e_c/ \sum_{c=1}^C e_c$, the hyper-parameter $\alpha$ is used to balance between the effective number and entropy. The convex function $\delta(\cdot)$ is utilized to map the sampling function better according to FlexMatch~\cite{zhang2021flexmatch}.
Finally, we can obtain a more comprehensive sampling function $\hat{s}_c$ for CamMix.

\noindent\textbf{Entropy-based Selection Module.}
Previous work~\cite{berthelot2019mixmatch, berthelot2019remixmatch, verma2022interpolation} in SSL primarily uses unlabeled samples for data mixing. 
Yet, some pseudo labels possess high uncertainty~\cite{arazo2020pseudo, wang2022freematch}, especially for challenging samples or in early training stages, causing confirmation bias~\cite{arazo2020pseudo}. 
Our data mixing approach allows the selection of both labeled and unlabeled data. 
We suggest augmenting high-uncertainty unlabeled data with confident labeled data.
However, this beneficial mixing may under-utilize unlabeled data when the regions of labeled data cover unlabeled ones, leaving them unexploited in training. This trade-off is a crucial consideration in data mixing.
Thus, we utilize sample-wise entropy $e_m$ as the selection indicator between labeled and unlabeled samples in data mixing as:
\begin{equation}
    e_m = \sum_{c=1}^{C}-f_c(A_w(u_m))\log(f_c(A_w(u_m))).
    \label{equ:8}
\end{equation}

We then define $M_{h}(\cdot)$ and $M_{l}(\cdot)$ as the masks of high and low entropy for selecting labeled and unlabeled samples respectively. They are also used to mask the unsupervised loss as in the Appendix~\textcolor{red}{A}. These masks can be expressed as:
\begin{equation}
\begin{aligned}
&M_{h}(u_m) = \mathbb{I}(e_m>\tau_e) \\
  &M_{l}(u_m)   =  \mathbb{I}(e_m<\tau_e),
  \label{equ:9}
\end{aligned}
\end{equation}
where $\tau_e$  is the selection threshold of the entropy mask, updated in EMA manner:
\begin{equation}
    \tau_e \leftarrow \lambda_\tau \tau_e + (1 - \lambda_\tau) e^t,
    \label{equ:10}
\end{equation}
where $\lambda_\tau$ denotes the EMA weight, $e^t$ is the average entropy of unlabeled data at each training iteration $t$, \emph{i.e.} $e^t = \frac{1}{B}\sum_{m=1}^{B} e_m$. In the early training stages, we select more labeled samples for mixing due to the uncertainty of model prediction on some unlabeled data. As training progresses and predictions become more reliable,  the utilization of unlabeled data increases.

\noindent\textbf{Entropy-based class balanced loss.} 
We further apply the class balanced loss, which is first introduced in~\cite{cui2019class} to re-balance the class distribution by utilizing the weighted loss based on the class-wise effective number as $L_{cb}=L_{cls}/E_c$. By normalizing $1/{E_c}$ as Eq.~\ref{equ:4}, we can obtain $L_{cb}=s_c L_{cls}$.

Moreover, to tackle the class-wise uncertainty problem in LTSSL, we propose entropy-based class balanced loss $L_{ecb}$ on the unlabeled data. $L_{ecb}$ uses $\hat{s}_c$ to measure both the effective number and uncertainty as:
\begin{equation}
    L_{ecb} =\hat{s}_c^u L_{cls},
    \label{equ:11}
\end{equation}
where $\hat{s}_c^u$ is calculated by Eq.~\ref{equ:alpha}, but only based on unlabeled data.
Finally, $L_{ecb}$ can be weighted towards both tail classes and high uncertainty classes, further re-balancing the training process.
Unlike previous entropy-based losses~\cite{grandvalet2004semi, saito2019semi}, our loss focuses on class-wise uncertainty instead of sample-wise, making it ideally suited for the LTSSL problem characterized by large category gaps.
A detailed description of the loss functions can be found in the Appendix~\textcolor{red}{A}.

\section{Experiments}

\subsection{Experimental setup}
\noindent\textbf{Datasets.}
We perform evaluation experiments of our proposed method on widely used long-tailed datasets, including CIFAR10-LT~\cite{krizhevsky2009learning}, CIFAR100-LT~\cite{krizhevsky2009learning}, STL10-LT~\cite{coates2011analysis} and ImageNet-127~\cite{fan2022cossl}.
To create imbalanced versions of the datasets, we randomly discard training samples to maintain the pre-defined imbalance ratio. With the imbalance ratio $\gamma_l$ and maximum number $N_1$ of labeled samples, we can calculate the number of labeled samples for class $c$ as $N_c = N_1 \times \gamma_l^{-\frac{c-1}{C-1}}$. Similarly, using the parameters $\gamma_u$ and $M_1$, we can determine the class distribution of unlabeled data quantity as in the labeled samples. 
For a detailed introduction to the datasets, please refer to the Appendix~\textcolor{red}{B}.

\noindent\textbf{Implementation Details.} Following DASO~\cite{oh2022daso}, we apply our method to various baseline frameworks, including FixMatch~\cite{sohn2020fixmatch}, FixMatch + LA~\cite{menon2020long}, FixMatch + ABC~\cite{lee2021abc} and  FixMatch + ACR~\cite{wei2023towards}. We compare our method with recent re-balancing methods like  DARP~\cite{kim2020distribution}, CReST/CReST+~\cite{wei2021crest} and DASO~\cite{oh2022daso}.
For a fair comparison, our code is developed based on DASO and ACR, implemented with Pytorch~\cite{paszke2019pytorch}.
We conduct our experiments  on CIFAR10-LT, CIFAR100-LT and STL-10 using Wide ResNet-28-2~\cite{zagoruyko2016wide}, and  on ImageNet-127 using ResNet-50~\cite{he2016deep}. The top-1 accuracy on the test set is used as the evaluation metric. The mean and standard deviation of three independent runs are reported. Due to the page limitation, detailed training settings are provided in the Appendix~\textcolor{red}{B}.

\subsection{Results on CIFAR10/100-LT and STL10-LT.}
We first consider the $\gamma_l = \gamma_u$ situation which is the most common scenario in SSL. Then, we investigate the performance of the methods by setting $\gamma_l \neq \gamma_u$, including uniform ($\gamma_u=1$) and reversed ($\gamma_u=1/100$) scenarios.

\noindent\textbf{In case of $\gamma_l = \gamma_u$.}
As shown in Tab.~\ref{tab:sota}, we compare our method with existing re-balancing methods under various baseline settings. When setting FixMatch as the baseline, our BEM shows superior performance improvement in most scenarios.  When further adding LA to FixMatch for label re-balancing,  our BEM outperforms all other configurations. When integrating ABC into FixMatch for pseudo label re-balancing, our BEM can benefit the baseline more than the DASO. Finally, we also demonstrate that our methods can complement ACR,  achieving the SOTA performance with an average gain of 18.35\% over FixMatch for CIFAR10-LT.
In summary, our BEM achieves consistent and significant gain under all baseline settings, showing its great adaptability.
The main reason is that, unlike most previous methods with pseudo label or logit adjustment, we directly re-balance the class distribution through data mixing, a vital technique missing in them, thus complementing these methods.

\begin{table*}[!htp]
\setlength\tabcolsep{3.5pt}
\begin{center}
\caption{Comparison of test accuracy with combinations of different baseline frameworks under $\gamma_l=\gamma_u$ setup on CIFAR10-LT and CIFAR100-LT. The best results for each diversion are in \textbf{bold}.}
\label{tab:sota}
\resizebox{0.95\textwidth}{!}{
\begin{tabular}{llcccccccc}
\toprule[1pt]
& & \multicolumn{4}{c}{CIFAR10-LT}& \multicolumn{4}{c}{CIFAR100-LT}\\
& & \multicolumn{2}{c}{$\gamma=\gamma_l=\gamma_u=100$}& \multicolumn{2}{c}{$\gamma=\gamma_l=\gamma_u=150$}&\multicolumn{2}{c}{$\gamma=\gamma_l=\gamma_u=10$}&\multicolumn{2}{c}{$\gamma=\gamma_l=\gamma_u=20$}\\
\cmidrule(r){3-4} \cmidrule(r){5-6} \cmidrule(r){7-8} \cmidrule(r){9-10}
& &$N_1=500$ &$N_1=1500$ &$N_1=500$ &$N_1=1500$ &$N_1=50$ &$N_1=150$ &$N_1=50$ &$N_1=150$ \\
Algorithm& &$M_1=4000$ &$M_1=3000$ &$M_1=4000$ &$M_1=3000$ &$M_1=400$ &$M_1=300$ &$M_1=400$ &$M_1=300$ \\
\cmidrule(r){1-2} \cmidrule(r){3-4} \cmidrule(r){5-6} \cmidrule(r){7-8} \cmidrule(r){9-10}
\multicolumn{2}{l}{Supervised}&47.3\footnotesize{±0.95} &61.9\footnotesize{±0.41} &44.2\footnotesize{±0.33} &58.2\footnotesize{±0.29} &29.6\footnotesize{±0.57} &46.9\footnotesize{±0.22} &25.1\footnotesize{±1.14} &41.2\footnotesize{±0.15}\\
\multicolumn{2}{l}{\quad w/LA~\cite{menon2020long}}&53.3\footnotesize{±0.44}&70.6\footnotesize{±0.21}&49.5\footnotesize{±0.40}&67.1\footnotesize{±0.78}&30.2\footnotesize{±0.44}&48.7\footnotesize{±0.89}&26.5\footnotesize{±1.31}&44.1\footnotesize{±0.42}\\
\cmidrule(r){1-2} \cmidrule(r){3-4} \cmidrule(r){5-6} \cmidrule(r){7-8} \cmidrule(r){9-10}

\multicolumn{2}{l}{FixMatch~\cite{sohn2020fixmatch}}&67.8\footnotesize{±1.13}&77.5\footnotesize{±1.32}&62.9\footnotesize{±0.36}&72.4\footnotesize{±1.03}&45.2\footnotesize{±0.55}&56.5\footnotesize{±0.06}&40.0\footnotesize{±0.96}&50.7\footnotesize{±0.25}\\
\multicolumn{2}{l}{\quad w/DARP~\cite{kim2020distribution}}&74.5\footnotesize{±0.78}&77.8\footnotesize{±0.63}&67.2\footnotesize{±0.32}&73.6\footnotesize{±0.73}&49.4\footnotesize{±0.20}&58.1\footnotesize{±0.44}&43.4\footnotesize{±0.87}&52.2\footnotesize{±0.66}\\
\multicolumn{2}{l}{\quad w/CReST+~\cite{wei2021crest}}&\textbf{76.3}\footnotesize{±0.86}&78.1\footnotesize{±0.42}&67.5\footnotesize{±0.45}&73.7\footnotesize{±0.34}&44.5\footnotesize{±0.94}&57.4\footnotesize{±0.18}&40.1\footnotesize{±1.28}&52.1\footnotesize{±0.21}
\\
\multicolumn{2}{l}{\quad w/DASO~\cite{oh2022daso}}&76.0\footnotesize{±0.37}&79.1\footnotesize{±0.75}&\textbf{70.1}\footnotesize{±1.81}&75.1\footnotesize{±0.77}&49.8\footnotesize{±0.24}&\textbf{59.2}\footnotesize{±0.35}&43.6\footnotesize{±0.09}&52.9\footnotesize{±0.42}\\
\multicolumn{2}{l}{\quad w/BEM (ours)}&75.8\footnotesize{±1.13}&\textbf{80.3}\footnotesize±0.62{}&69.7\footnotesize{±0.91}&\textbf{75.7}\footnotesize{±0.22}&\textbf{50.4}\footnotesize{±0.34}&59.0\footnotesize{±0.23}&\textbf{44.1}\footnotesize{±0.18}&\textbf{54.3}\footnotesize{±0.36}\\
\cmidrule(r){1-2} \cmidrule(r){3-4} \cmidrule(r){5-6} \cmidrule(r){7-8} \cmidrule(r){9-10}

\multicolumn{2}{l}{FixMatch+LA~\cite{menon2020long}}&75.3\footnotesize{±2.45}&82.0\footnotesize{±0.36}&67.0\footnotesize{±2.49}&78.0\footnotesize{±0.91}&47.3\footnotesize{±0.42}&58.6\footnotesize{±0.36}&41.4\footnotesize{±0.93}&53.4\footnotesize{±0.32}\\

\multicolumn{2}{l}{\quad w/DARP~\cite{kim2020distribution}}&76.6\footnotesize{±0.92}&80.8\footnotesize{±0.62}&68.3\footnotesize{±0.94}&76.7\footnotesize{±1.13}&50.5\footnotesize{±0.78}&59.9\footnotesize{±0.32}&44.4\footnotesize{±0.65}&53.8\footnotesize{±0.43}\\

\multicolumn{2}{l}{\quad w/CReST~\cite{wei2021crest}}&76.7\footnotesize{±1.13}&81.1\footnotesize{±0.57}&70.9\footnotesize{±1.18}&77.9\footnotesize{±0.71}&44.0\footnotesize{±0.21}&57.1\footnotesize{±0.55}&40.6\footnotesize{±0.55}&52.3\footnotesize{±0.20}\\

\multicolumn{2}{l}{\quad w/DASO~\cite{oh2022daso}}&77.9\footnotesize{±0.88}&82.5\footnotesize{±0.08}&70.1\footnotesize{±1.68}&79.0\footnotesize{±2.23}&50.7\footnotesize{±0.51}&60.6\footnotesize{±0.71}&44.1\footnotesize{±0.61}&55.1\footnotesize{±0.72}\\
\multicolumn{2}{l}{\quad w/BEM (ours)}&\textbf{78.6}\footnotesize{±0.97}&\textbf{83.1}\footnotesize{±0.13}&\textbf{72.5}\footnotesize{±1.13}&\textbf{79.9}\footnotesize{±1.02}&\textbf{51.3}\footnotesize{±0.26}&\textbf{61.9}\footnotesize{±0.57}&\textbf{44.8}\footnotesize{±0.21}&\textbf{56.1}\footnotesize{±0.54}\\

\cmidrule(r){1-2} \cmidrule(r){3-4} \cmidrule(r){5-6} \cmidrule(r){7-8} \cmidrule(r){9-10}
\multicolumn{2}{l}{FixMatch+ABC~\cite{lee2021abc}}&78.9\footnotesize{±0.82}&83.8\footnotesize{±0.36}&66.5\footnotesize{±0.78}&80.1\footnotesize{±0.45}&47.5\footnotesize{±0.18}&59.1\footnotesize{±0.21}&41.6\footnotesize{±0.83}&53.7\footnotesize{±0.55}\\

\multicolumn{2}{l}{\quad w/DASO~\cite{oh2022daso}}&\textbf{80.1}\footnotesize{±1.16}&83.4\footnotesize{±0.31}&70.6\footnotesize{±0.80}&80.4\footnotesize{±0.56}&\textbf{50.2}\footnotesize{±0.62}&60.0\footnotesize{±0.32}&\textbf{44.5}\footnotesize{±0.25}&55.3\footnotesize{±0.53}\\
\multicolumn{2}{l}{\quad w/BEM (ours)}&79.8\footnotesize{±0.82}&\textbf{83.9}\footnotesize{±0.34}&\textbf{70.7}\footnotesize{±0.78}&\textbf{80.8}\footnotesize{±0.67}&50.0\footnotesize{±0.15}&\textbf{60.9}\footnotesize{±0.42}&44.4\footnotesize{±0.18}&\textbf{55.5}\footnotesize{±0.84}\\

\cmidrule(r){1-2} \cmidrule(r){3-4} \cmidrule(r){5-6} \cmidrule(r){7-8} \cmidrule(r){9-10}
\multicolumn{2}{l}{FixMatch+ACR~\cite{wei2023towards}}&81.6\footnotesize{±0.19}&84.1\footnotesize{±0.39}&77.0\footnotesize{±1.19}&80.9\footnotesize{±0.22}&55.7\footnotesize{±0.12}&65.6\footnotesize{±0.16}&48.0\footnotesize{±0.75}&58.9\footnotesize{±0.36}\\
\multicolumn{2}{l}{\quad w/BEM (ours)}&\textbf{83.5}\footnotesize{±0.33}&\textbf{85.5}\footnotesize{±0.28}&\textbf{78.1}\footnotesize{±0.99}&\textbf{83.8}\footnotesize{±1.12}&\textbf{55.8}\footnotesize{±0.32}&\textbf{66.3}\footnotesize{±0.24}&\textbf{48.6}\footnotesize{±0.45}&\textbf{59.8}\footnotesize{±0.37}\\
\toprule[1pt]
\end{tabular}
}
\end{center}
\vspace{-10pt}
\end{table*}

\begin{table*}[!htp]
\setlength\tabcolsep{3.5pt}
\begin{center}
\caption{Comparison of test accuracy with combinations of different baseline frameworks under $\gamma_l \neq \gamma_u$ setup on CIFAR10-LT and STL10-LT. The $\gamma_l$ is fixed to 100 for CIFAR10-LT, and the $\gamma_l$ is set to 10 and 20 for STL10-LT. The $N/A$ denotes the class distribution of data quantity is unknown. The best results for each diversion are in \textbf{bold}.}
\label{tab:sota2}
\resizebox{0.95\textwidth}{!}{
\begin{tabular}{llcccccccc}
\toprule[1pt]
& & \multicolumn{4}{c}{CIFAR10-LT($\gamma_l \neq \gamma_u$)}& \multicolumn{4}{c}{STL10-LT($\gamma_u = N/A$)}\\
& & \multicolumn{2}{c}{$\gamma_u=1$(uniform)}&\multicolumn{2}{c}{$\gamma_u=1/100$(reversed)}&\multicolumn{2}{c}{$\gamma_l=10$}& \multicolumn{2}{c}{$\gamma_l=20$}\\
\cmidrule(r){3-4} \cmidrule(r){5-6} \cmidrule(r){7-8} \cmidrule(r){9-10}
& &$N_1=500$ &$N_1=1500$ &$N_1=500$ &$N_1=1500$ &$N_1=150$ &$N_1=450$ &$N_1=150$ &$N_1=450$ \\
Algorithm& &$M_1=4000$ &$M_1=3000$ &$M_1=4000$ &$M_1=3000$ &$M=100k$ &$M=100k$ &$M=100k$ &$M=100k$ \\
\cmidrule(r){1-2} \cmidrule(r){3-4} \cmidrule(r){5-6} \cmidrule(r){7-8} \cmidrule(r){9-10}
\multicolumn{2}{l}{FixMatch~\cite{sohn2020fixmatch}}&73.0\footnotesize{±3.81}& 81.5\footnotesize{±1.15}& 62.5\footnotesize{±0.94}& 71.8\footnotesize{±1.70}& 56.1\footnotesize{±2.32}& 72.4\footnotesize{±0.71}& 47.6\footnotesize{±4.87}& 64.0\footnotesize{±2.27}\\
\multicolumn{2}{l}{\quad w/DARP~\cite{kim2020distribution}}&82.5\footnotesize{±0.75}& 84.6\footnotesize{±0.34}& 70.1\footnotesize{±0.22}& 80.0\footnotesize{±0.93}& 66.9\footnotesize{±1.66}& 75.6\footnotesize{±0.45}& 59.9\footnotesize{±2.17}& 72.3\footnotesize{±0.60}\\
\multicolumn{2}{l}{\quad w/CReST~\cite{wei2021crest}}&83.2\footnotesize{±1.67}& 87.1\footnotesize{±0.28}& 70.7\footnotesize{±2.02}& \textbf{80.8}\footnotesize{±0.39} &61.7\footnotesize{±2.51} &71.6\footnotesize{±1.17} & 57.1\footnotesize{±3.67} &68.6\footnotesize{±0.88} \\
\multicolumn{2}{l}{\quad w/CReST+~\cite{wei2021crest}}&82.2\footnotesize{±1.53}& 86.4\footnotesize{±0.42}& 62.9\footnotesize{±1.39}& 72.9\footnotesize{±2.00}& 61.2\footnotesize{±1.27}& 71.5\footnotesize{±0.96}& 56.0\footnotesize{±3.19}& 68.5\footnotesize{±1.88}\\
\multicolumn{2}{l}{\quad w/DASO~\cite{oh2022daso}}&86.6\footnotesize{±0.84}& 88.8\footnotesize{±0.59} &\textbf{71.0}\footnotesize{±0.95} &80.3\footnotesize{±0.65}& \textbf{70.0}\footnotesize{±1.19}&78.4\footnotesize{±0.80}& \textbf{65.7}\footnotesize{±1.78} &75.3\footnotesize{±0.44}\\
\multicolumn{2}{l}{\quad w/BEM(ours)}&\textbf{86.8}\footnotesize{±0.47}&\textbf{89.1}\footnotesize{±0.75}&70.0\footnotesize{±1.72}&79.1\footnotesize{±0.77}&
68.3\footnotesize{±1.15}&\textbf{81.2}\footnotesize{±1.42}&61.6\footnotesize{±0.98} &\textbf{76.0}\footnotesize{±1.51}\\
\cmidrule(r){1-2} \cmidrule(r){3-4} \cmidrule(r){5-6} \cmidrule(r){7-8} \cmidrule(r){9-10}
\multicolumn{2}{l}{FixMatch+ACR~\cite{wei2023towards}}&92.1\footnotesize{±0.18}&93.5\footnotesize{±0.11}&85.0\footnotesize{±0.09}&89.5\footnotesize{±0.17}&77.1\footnotesize{±0.24}&83.0\footnotesize{±0.32}&75.1\footnotesize{±0.70}&81.5\footnotesize{±0.25}\\
\multicolumn{2}{l}{\quad w/BEM(ours)}&\textbf{94.3}\footnotesize{±0.14}&\textbf{95.1}\footnotesize{±0.56}&\textbf{85.5}\footnotesize{±0.21}&\textbf{89.8}\footnotesize{±0.12}&\textbf{79.3}\footnotesize{±0.34}&\textbf{84.2}\footnotesize{±0.56}&\textbf{75.9}\footnotesize{±0.15}&\textbf{82.3}\footnotesize{±0.23}\\
\toprule[1pt]
\end{tabular}
}
\end{center}
\vspace{-10pt}
\end{table*}

\noindent\textbf{In case of $\gamma_l  \neq \gamma_u$.} In real-world datasets, the class distribution of unlabeled data remains unknown or inconsistent with labeled data. For CIFAR10-LT, we consider two extreme scenarios: uniform and reversed. For STL10-LT, where the class distribution of unlabeled data is unknown, we set $\gamma_l\in\{10,20\}$ and $N _1 \in\{150, 450\}$. 

As shown in Tab.~\ref{tab:sota2}, our methods yield an average improvement of 14.1\% and 11.1\% over FixMatch in two scenarios for CIFAR10-LT.  However, our method is less effective than DASO under the reversed setting. 
We speculate that data mixing methods cannot achieve thorough re-balancing in challenging scenarios, unlike approaches from the prediction perspective.
However, integrating ACR results in the best performance on CIFAR10-LT, even under the reversed setting, with an average gain of 22.9\% and 30.9\% over FixMatch. Similarly, for STL-10, our method enhances the performance of FixMatch and achieves the best performance when combined with  ACR.
This highlights the value of our BEM for re-balancing methods. 
Further comparisons of our method with more re-balancing methods can be found in the Appendix~\textcolor{red}{D}.

\noindent\textbf{BEM on the SSL learner.} 
We further validate the adaptability of BEM with various SSL learners, including MeanTeacher~\cite{tarvainen2017mean}, FlexMatch~\cite{zhang2021flexmatch} and SoftMatch~\cite{chen2023softmatch}. Notably, FlexMatch and SoftMatch outperform FixMatch on balanced datasets. For SoftMatch, we only apply $L_{cb}$  considering its training process already re-weights the loss based on class-wise confidence.
Following DASO, we set $\gamma_l=100$ for CIFAR10-LT and $\gamma_l=10$ for CIFAR100-LT and STL10-LT. 
As depicted in Tab.~\ref{tab:SSL}, our method enhances the performance of all SSL learners under each setting. Specially, MeanTeacher initially underperforms on the Long-Tailed dataset but achieves gains of 41.1\%, 15.4\%, and 37.9\% on three datasets by applying BEM. SoftMatch, the state-of-the-art SSL method, also gains an additional 5.5\%, 9.7\% and 9.7\% improvement with our BEM.

\begin{table}[!tp]
\setlength\tabcolsep{2pt}
\begin{center}
\caption{Comparison of test accuracy with combinations of different SSL learners, including MeanTeacher, FlexMatch and SoftMatch.}
\label{tab:SSL}
\resizebox{0.45\textwidth}{!}{
\begin{tabular}{llcccc}
\toprule[1pt]
& & \multicolumn{2}{c}{C10-LT}&  \multicolumn{1}{c}{C100-LT}&  \multicolumn{1}{c}{STL10-LT}\\
\cmidrule(r){3-4} \cmidrule(r){5-5}  \cmidrule(r){6-6} 
& & \multicolumn{2}{c}{$N_1=1500$}&  \multicolumn{1}{c}{$N_1=150$}&  \multicolumn{1}{c}{$N_1=450$}\\
& & \multicolumn{2}{c}{$M_1=3000$}&  \multicolumn{1}{c}{$M_1=300$}&  \multicolumn{1}{c}{$M=100k$}\\
\cmidrule(r){3-4} \cmidrule(r){5-5}  \cmidrule(r){6-6} 
Algorithm& &$\gamma_u=100$&$\gamma_u=1$&$\gamma_u=10$&$\gamma_u=N/A$\\
\cmidrule(r){1-2} \cmidrule(r){3-4} \cmidrule(r){5-5}  \cmidrule(r){6-6}
\multicolumn{2}{l}{MeanTeacher\cite{tarvainen2017mean}}&  68.6\footnotesize{±0.88} & 46.4\footnotesize{±0.98} & 52.1\footnotesize{±0.09} & 54.6\footnotesize{±0.17} \\
\multicolumn{2}{l}{\quad w/BEM(Ours)}&  \textbf{73.5}\footnotesize{±0.56} & \textbf{81.3}\footnotesize{±1.67} & \textbf{60.1}\footnotesize{±0.43} & \textbf{75.3}\footnotesize{±0.59} \\
\cmidrule(r){1-2} \cmidrule(r){3-4} \cmidrule(r){5-5}  \cmidrule(r){6-6}
\multicolumn{2}{l}{FlexMatch~\cite{zhang2021flexmatch}}&  79.2\footnotesize{±0.92}& 82.2\footnotesize{±0.23} & 62.1\footnotesize{±0.86} & 74.9\footnotesize{±0.42} \\
\multicolumn{2}{l}{\quad w/BEM(Ours)}& \textbf{81.2}\footnotesize{±0.50} & \textbf{88.0}\footnotesize{±0.17} & \textbf{68.4}\footnotesize{±0.79} & \textbf{81.2}\footnotesize{±0.92} \\
\cmidrule(r){1-2} \cmidrule(r){3-4} \cmidrule(r){5-5}  \cmidrule(r){6-6}
\multicolumn{2}{l}{SoftMatch~\cite{chen2023softmatch}}&  79.6\footnotesize{±0.46} & 78.3\footnotesize{±0.86} & 62.8\footnotesize{±0.33}& 75.5\footnotesize{±0.74} \\
\multicolumn{2}{l}{\quad w/BEM(Ours)}&  \textbf{82.0}\footnotesize{±0.38} & \textbf{84.5}\footnotesize{±0.25} & \textbf{68.9}\footnotesize{±1.08} & \textbf{82.8}\footnotesize{±0.49} \\
\toprule[1pt]
\end{tabular}
}
\end{center}
\vspace{-20pt}
\end{table}

\subsection{Results on ImageNet-127.}
ImageNet127, initially introduced in \cite{huh2016makes} and later employed by CReST \cite{wei2021crest} for imbalanced SSL, is a naturally imbalanced dataset with an imbalance ratio $\gamma 	\approx 286$. It groups the 1000 classes of ImageNet~\cite{deng2009imagenet} into 127 classes, based on the WordNet hierarchy. Due to resource constraints, 
we down-sample the origin ImageNet127 images to $32\times32$ or $64\times64$ pixel images~\cite{fan2022cossl} and randomly select 10\% of training samples as the labeled set. Given the long-tailed test set, we set $\alpha=0.2$ to reduce sampling and loss weight bias towards tail classes, favoring high uncertainty classes instead. Tab.~\ref{tab:imagenet} demonstrates the superiority of our method over FixMatch, even without other re-balancing techniques. When combined with ACR, our method achieves the best results for both image sizes  (95.3\% and 51.1\% absolute gains over FixMatch).  This shows the applicability of our BEM  to long-tailed test datasets 
and its ability to enhance previous re-balancing methods.

\begin{table}[!t]
\begin{center}
\caption{Comparison of test accuracy with combinations of different baseline frameworks on ImageNet-127. 
}
\label{tab:imagenet}
\resizebox{0.4\textwidth}{!}{
\begin{tabular}{llcc}
\toprule[1pt]
Algorithm & &$32\times32$&$64\times64$\\
\cmidrule(r){1-2} \cmidrule(r){3-3} \cmidrule(r){4-4}
\multicolumn{2}{l}{FixMatch~\cite{sohn2020fixmatch}}&29.7&42.3\\
\multicolumn{2}{l}{\quad w/DARP~\cite{kim2020distribution}}&30.5& 42.5 \\
\multicolumn{2}{l}{\quad w/DARP+cRT~\cite{kim2020distribution}}&39.7 &51.0 \\
\multicolumn{2}{l}{\quad w/CReST+~\cite{wei2021crest}}&32.5 &44.7 \\
\multicolumn{2}{l}{\quad w/CReST++LA~\cite{menon2020long}}&40.9 &55.9 \\
\multicolumn{2}{l}{\quad w/CoSSL~\cite{fan2022cossl}}&43.7 &53.9 \\
\multicolumn{2}{l}{\quad w/TRAS~\cite{wei2022transfer}}&46.2 &54.1 \\
\multicolumn{2}{l}{\quad w/BEM(Ours)}&53.3 &58.2 \\
\multicolumn{2}{l}{\quad w/ACR~\cite{wei2023towards}}&57.2 &63.6 \\
\multicolumn{2}{l}{\quad w/ACR+BEM(Ours)}&\textbf{58.0} &\textbf{63.9} \\
\toprule[1pt]
\end{tabular}
}
\end{center}
\vspace{-20pt}
\end{table}

\subsection{Comprehensive analysis of the method.}
We perform comprehensive ablation studies to further understand how our method enhances baseline frameworks. Following DASO, we use CIFAR10-LT (C10) with $N_1=500$, $\gamma=100$  and STL10-LT (STL10) with $N_1=150$, $\gamma_l=10$ to cover  both $\gamma_l=\gamma_u$ and  $\gamma_l \neq \gamma_u $ cases.  Our baseline framework is FixMatch. More results
are provided in the Appendix~\textcolor{red}{D}.

\noindent\textbf{Ablation study on different mixing strategies.}
We compare our mixing method with existing techniques including MixUp~\cite{zhang2017mixup}, CutMix~\cite{yun2019cutmix} and SaliencyMix~\cite{uddin2020saliencymix} to demonstrate its effectiveness in Tab.~\ref{tab:aug}. First, we mix data within the same batch. SaliencyMix outperforms CutMix and MixUp on both datasets, and our CamMix surpasses SaliencyMix, indicating better localization ability. 
By further optimizing the in-batch mixing method, BEM achieves the best results.
\begin{table}[!t]
\begin{center}
\caption{Ablation study on different mixing strategies. Apart from BEM, all other methods perform mixing within the same batch.}
\label{tab:aug}
\begin{tabular}{llcc}
\toprule[1pt]
Algorithm& &\multicolumn{1}{c}{C10}&  \multicolumn{1}{c}{STL10}\\
\hline
\multicolumn{2}{l}{FixMatch~\cite{sohn2020fixmatch}}&  67.8 & 56.1 \\
\multicolumn{2}{l}{\quad w/MixUp~\cite{zhang2017mixup}}&  69.9 & 63.2 \\
\multicolumn{2}{l}{\quad w/CutMix~\cite{yun2019cutmix}}&70.2 & 62.5 \\
\multicolumn{2}{l}{\quad w/SaliencyMix~\cite{uddin2020saliencymix}}&70.8 & 64.0 \\
\multicolumn{2}{l}{\quad w/CamMix(Ours)}&71.9 & 64.8 \\
\multicolumn{2}{l}{\quad w/BEM(Ours)}&\textbf{75.7} & \textbf{68.3} \\
\toprule[1pt]
\end{tabular}
\end{center}
\vspace{-20pt}
\end{table}

\noindent\textbf{Ablation study on each component of BEM.} 
We verify each component in BEM by either removal or standard component replacement in Tab.~\ref{tab:comp}.
The accuracy on both datasets reduces sharply when replacing CamMix with CutMix. It highlights the importance of semantic region selection.  We then remove CBMB and implement random sampling, resulting in a maximum performance decrease of 5.1\% and 4.9\%, respectively. This suggests our CBMB effectively tackles the long-tail problem.
Removing ESS, denoted as setting $\alpha=1$, also leads to a decline in the model's performance.
When we remove ESM and merely use unlabeled data mixing, it results in a 4.4\% performance decrease on STL10. This implies initial training phase guidance from confident labeled data 
resolves the problem of pseudo label errors especially when $\gamma_l \neq \gamma_u$. Finally, the removal of the ECB loss also causes a performance drop on both datasets.
\begin{table}[!t]
\begin{center}
\caption{Ablation study on each component of BEM.}
\label{tab:comp}
\begin{tabular}{lcc}
\toprule[1pt]
& \multicolumn{1}{c}{C10}&  \multicolumn{1}{c}{STL10}\\
\hline
\multicolumn{1}{l}{BEM(Ours)}&  \textbf{75.7} & \textbf{68.3} \\
\multicolumn{1}{l}{\quad w/o CamMix}&  74.0 & 66.6 \\
\multicolumn{1}{l}{\quad w/o CBMB}&  72.1 & 65.0 \\
\multicolumn{1}{l}{\quad w/o ESS}&  74.7 &67.0\\
\multicolumn{1}{l}{\quad w/o ESM}&  75.3 & 65.3 \\
\multicolumn{1}{l}{\quad w/o ECB Loss}&  74.9 & 67.2 \\
\toprule[1pt]
\end{tabular}
\end{center}
\vspace{-20pt}
\end{table}



        Furthermore,  we conduct a qualitative analysis of the performance enhancement achieved by BEM on CIFAR10-LT, setting $\gamma_l =\gamma_u =  100$, $N_1 = 500$ and $M_1 = 4000$. More visualization analysis can be seen in the Appendix~\textcolor{red}{E}.

\begin{figure}[!t]
    \centering
    \includegraphics[width=0.95\linewidth]{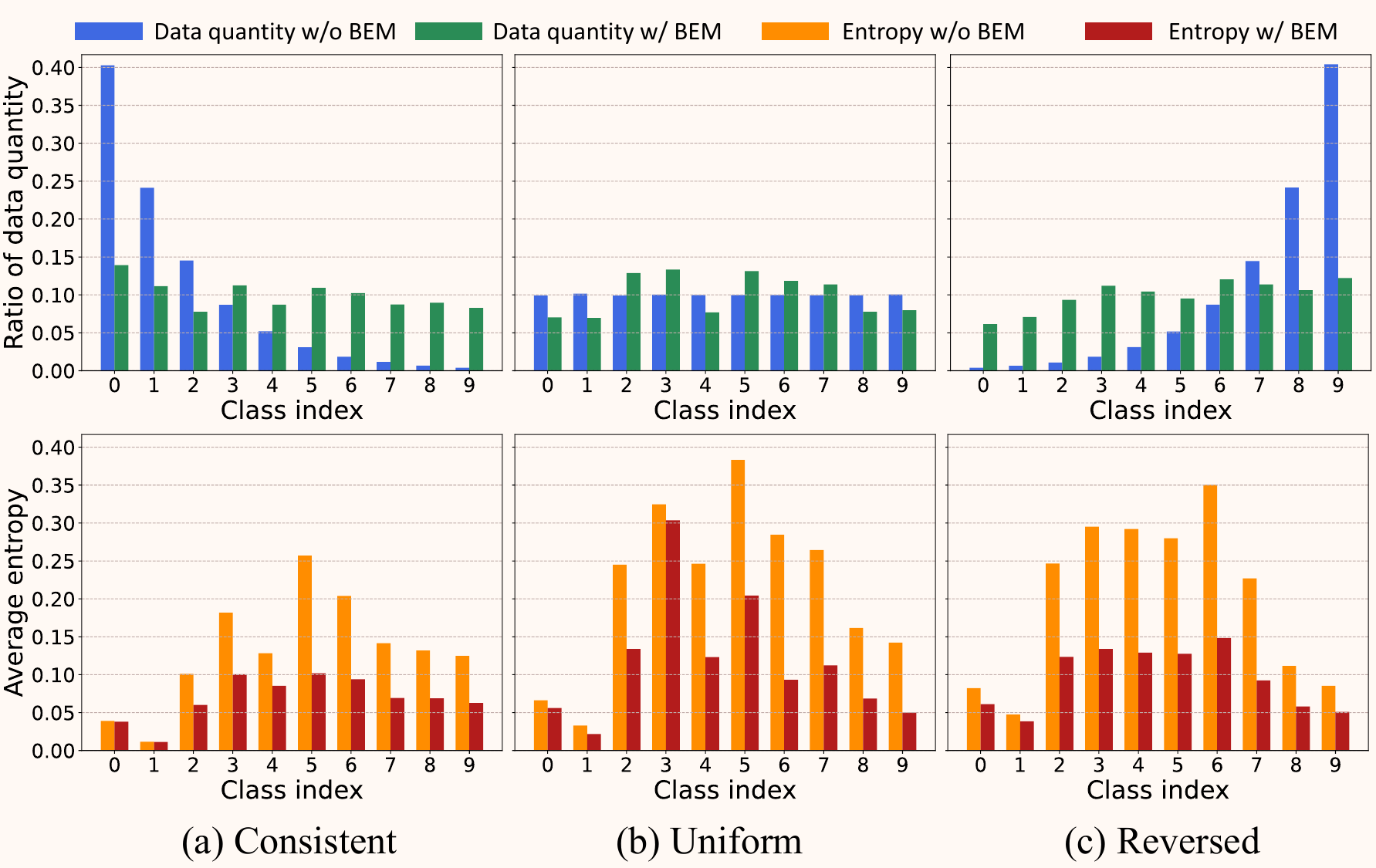}
    \caption{Class distribution of data quantity and entropy in three settings. 
    Each mixed data is calculated as containing two classes.
   }
    \label{fig:number_entropy}
    \vspace{-10pt}
\end{figure}

\noindent\textbf{Visualization of the class distribution of  unlabeled data quantity and entropy.} 
To verify the effect of BEM on the re-balancing training process, we visualize the class distribution of data quantity and entropy. 
Fig.~\ref{fig:number_entropy} reveals our method's effect on re-balancing data quantity across all settings. Moreover, our approach notably diminishes uncertainty via entropy and re-balances class-wise entropy, particularly in uniform settings where higher entropy classes engage more training samples, thus lowering uncertainty.

\begin{figure}[!t]
\centering
\subfloat[FixMatch]{\includegraphics[scale=0.3]{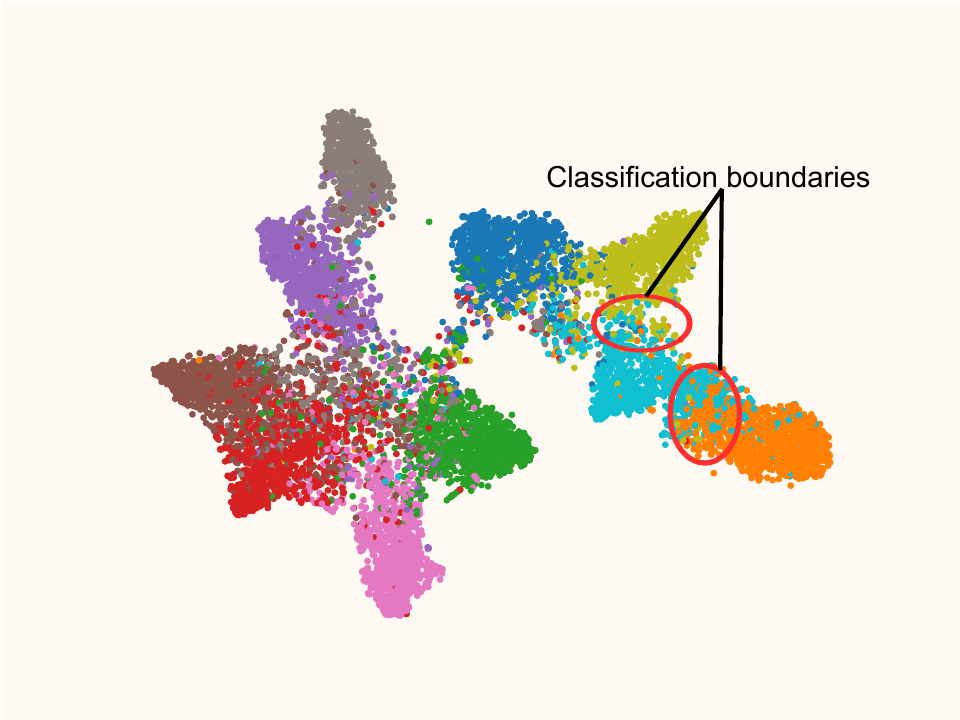}}
\subfloat[FixMatch w/BEM]{\includegraphics[scale=0.3]{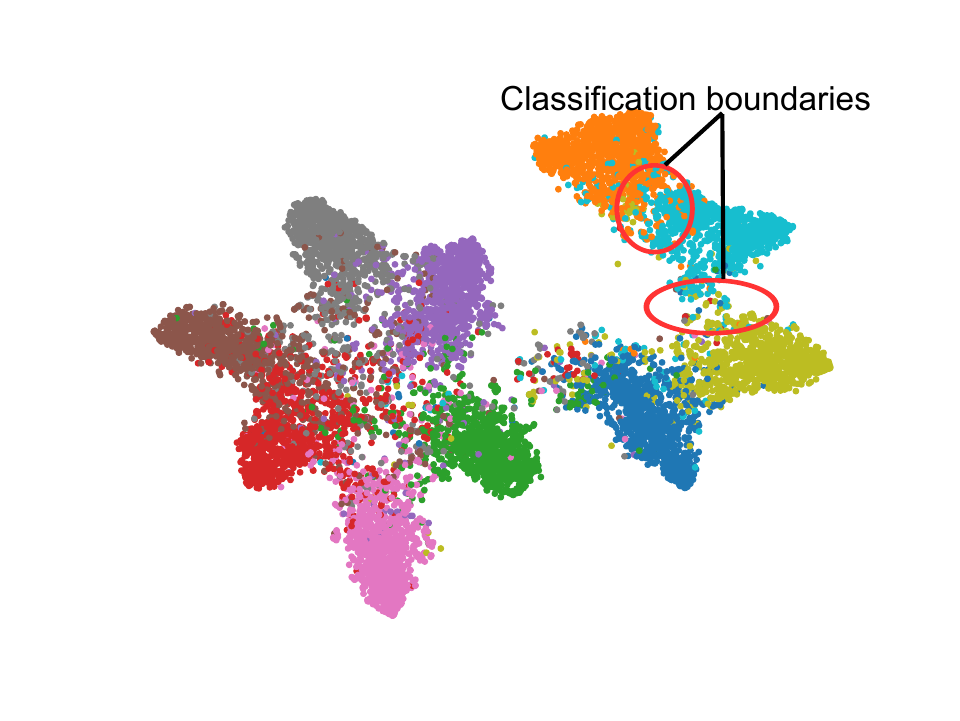}}\\
\subfloat[ACR]{\includegraphics[scale=0.3]{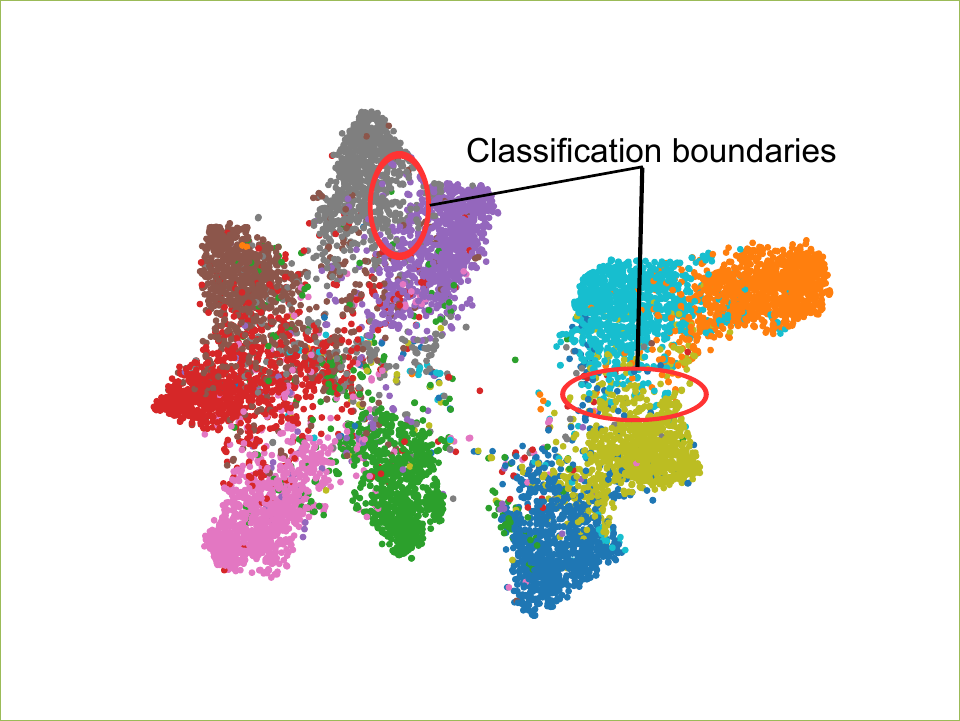}}
\subfloat[ACR w/BEM]{\includegraphics[scale=0.3]{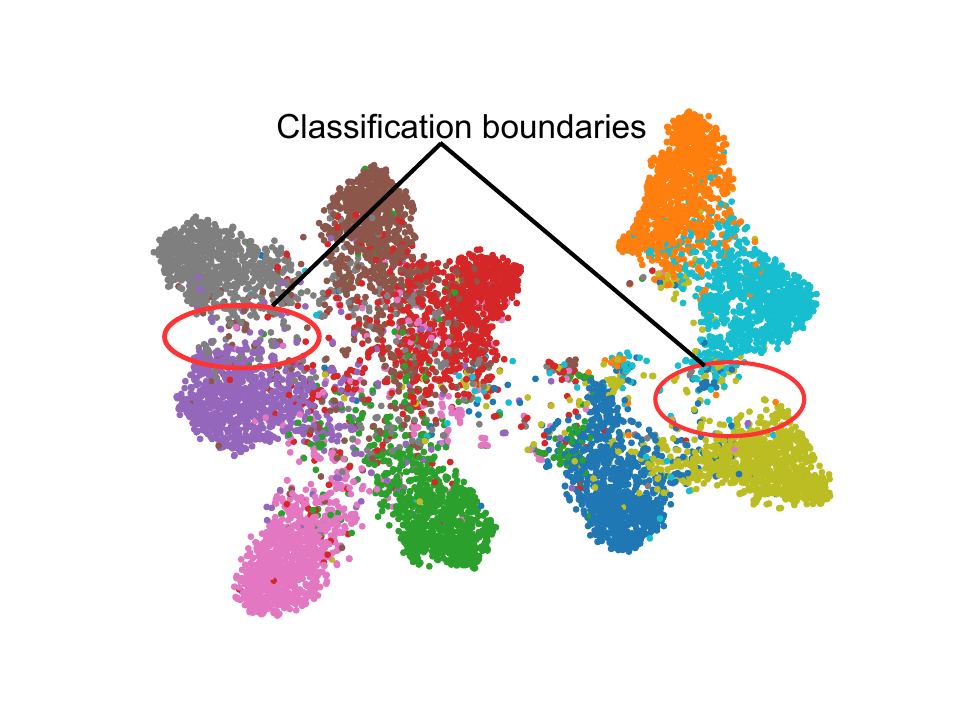}}\\
\caption{Comparison of t-SNE visualization with combinations of FixMatch and ACR.}
\label{fig:tsne}
\vspace{-10pt}
\end{figure}

\noindent\textbf{Visualization of T-SNE.} Additionally, we visualize the learning representation on the balanced test set using t-distributed stochastic neighbor embedding (t-SNE)~\cite{van2008visualizing}. We apply our method to FixMatch and ACR respectively. The results in Fig.~\ref{fig:tsne} suggest that our method generates clearer classification boundaries for representations.

\begin{figure}[!t]
    \centering
    \includegraphics[width=0.98\linewidth]{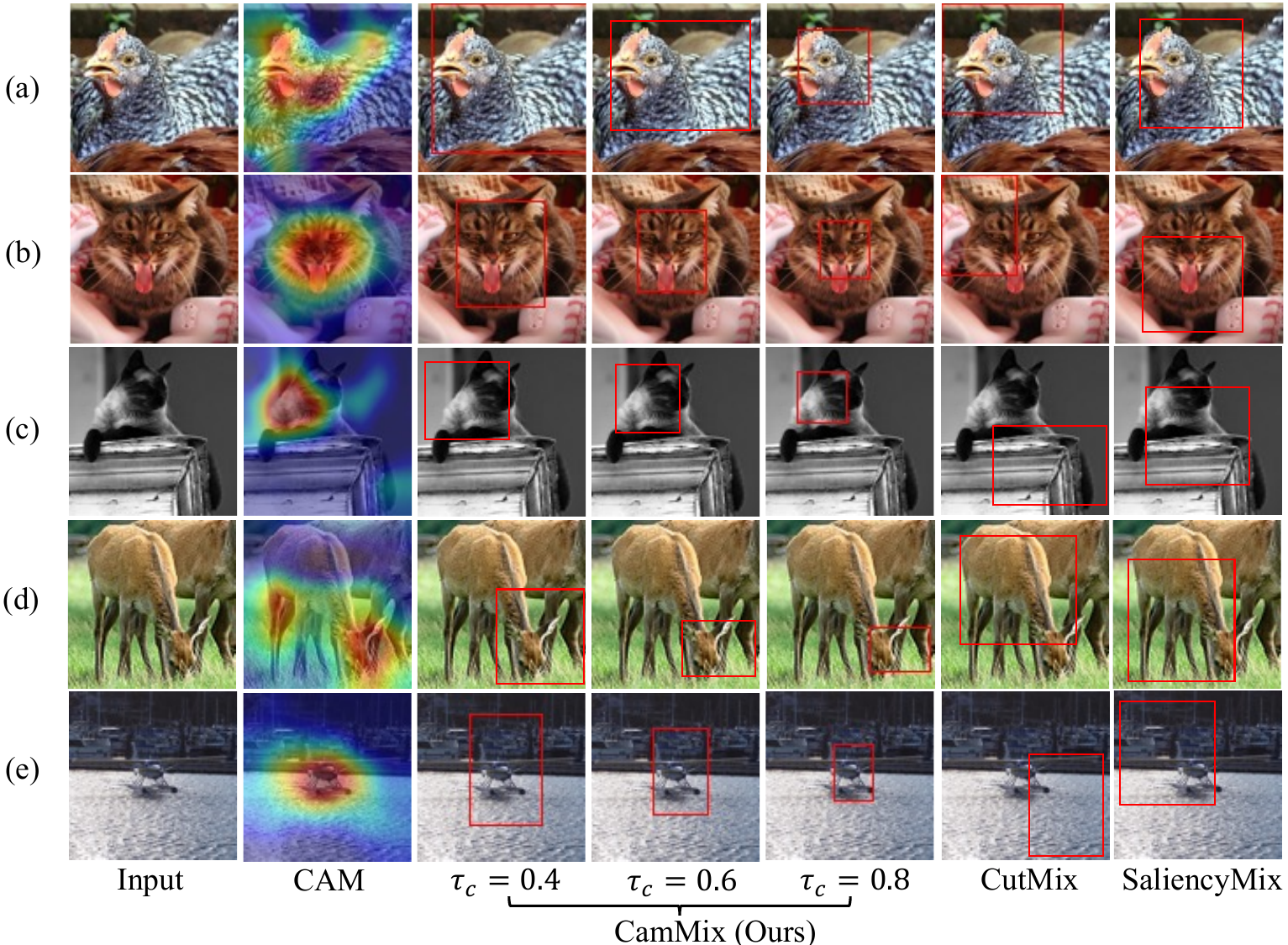}
    \caption{The visualization of data mixing process for CutMix, SaliencyMix, and CamMix on STL10-LT. The red box indicates the image area selected by data mixing.
   }
    \label{fig:vis mix1}
    \vspace{-10pt}
\end{figure}

\noindent\textbf{Visualization of data mixing.}
As shown in Fig.~\ref{fig:vis mix1}, we compare intermediate images from various data mixing methods on STL10 due to the high-resolution input. Five images with different target sizes are selected to visualize. CutMix shows strong randomness and tends to miss the class content, especially when the target is small (see in  (e)). Although SaliencyMix has initial target localization ability, it often fails to accurately locate key areas and tends to include numerous redundant contexts (see in (c) and (e)). CamMix shows the best localization ability, accurately locating the class content based on CAM. As $\tau_c$ increases, localization accuracy improves and inclusion of redundant context decreases.

\section{Conclusion}
 In this work, we introduce a novel approach, Balanced and Entropy-based Mix (BEM), to enhance long-tailed semi-supervised learning by re-balancing the training process. Specially, we re-balance data quantity using the class balanced mix bank and re-balance class-wise uncertainty through the entropy-based learning approach. As the first method to leverage data mixing in LTSSL, BEM significantly boosts the accuracy of various LTSSL frameworks across multiple benchmarks, offering a complementary technique for other re-balancing methods.

{\small
\bibliographystyle{ieeenat_fullname}
\bibliography{_main}
}

\ifarxiv \clearpage \appendix 




\maketitlesupplementary

\section{Detailed Loss Functions}
We detail the loss functions for the training in this section. For the labeled data, we directly adopt the cross entropy $\mathcal{H}(\cdot)$ to calculate the supervised loss $L_s$. For the unlabeled data, we first follow FixMatch~\cite{sohn2020fixmatch} to filter samples with low-confidence pseudo label by a mask $M_u(u_m) = \mathbb{I}(\max(f(A_w(u_m)))>\tau)$.
Then, we can obtain $M_h$ and $M_l$, the masks of high and low entropy for selecting labeled samples $(x_m^s, y_m^s)$ and unlabeled samples $u_m^s$ in data mixing. Given the mixed samples $u_m^\prime$ from CAMmix, we can obtain four types of unsupervised loss (\emph{i.e.} $L_u^h$, $L_{u}^{l},L_{u^s}^{h}$, and $L_{u^s}^{l}$), in which $L_u^h$, $L_{u}^{l}$ and $L_{u^s}^{l}$ are weighted by the entropy-based class balanced weight $\hat{s}^u$ to form the  $L_{ecb}$. Specifically, 
$L_u^h$ and $L_u^l$ are supervised by the pseudo label of the original unlabeled data $q_m$, while $L_{u^s}^h$ and $L_{u^s}^l$ are supervised by the ground truth of the sampled labeled data $y_m^s$ and the pseudo label of the sampled unlabeled data $q_m^s$, respectively. 
The final loss function $L$ is weighted by $\lambda$ reflecting the proportion of area occupied by original and sampled data as in CutMix~\cite{yun2019cutmix}. Detailed loss functions are as follows:
\begin{equation}
\begin{aligned}
L_{s}&= \sum_{n=1}^{B} \mathcal{H}(f(A_w(x_n)), y_n)     \\
L_{u}^{h} &= \hat{s}^u \sum_{m=1}^{B} M_u(u_m) M_{h}(u_m) \mathcal{H}(f(u_m^\prime), q_m)      \\
L_{u}^{l} &= \hat{s}^u \sum_{m=1}^{B} M_u(u_m) M_{l}(u_m) \mathcal{H}(f(u_m^\prime), q_m)      \\
L_{u^s}^{h}&= \sum_{m=1}^{B} M_{h}(u_m) \mathcal{H}(f(u_m^\prime), y_m^{s})    \\
L_{u^s}^{l} &= \hat{s}^u\sum_{m=1}^{B} M_u(u_m^{s}) M_{l}(u_m) \mathcal{H}(f(u_m^\prime), q_m^s)     \\
L &= L_{s} + \lambda( L_{u}^{h}+L_{u}^{l})+ (1-\lambda)(L_{u^s}^{h}+L_{u^s}^{l}) 
  \label{equ:loss}
\end{aligned}
\end{equation}

 %

\section{Detailed Experimental Setup}
\label{sec:1_setup}
In this section, we provide additional information about the datasets and implementation details.

\noindent\textbf{Datasets.} 
We evaluate our method in three scenarios, i.e., \emph{1)} the class distribution of labeled data is consistent with the unlabeled data ($\gamma_l = \gamma_u$). \emph{2)} the labeled and unlabeled data fail to share the same distribution ($\gamma_l \neq \gamma_u$). \emph{3)} The test data possesses an imbalanced class distribution.
\renewcommand{\algorithmicrequire}{\textbf{Input:}}
\renewcommand{\algorithmicensure}{\textbf{Require:}}
\begin{algorithm}[h]
\caption{Balanced and Entropy-based Mix (BEM)}
\label{alg:2_BEM}
\begin{algorithmic}[1]
\REQUIRE Labeled dataset $X$, unlabeled dataset $U$, model $f$,
effective number of labeled data $E_c^x$, 
CAM threshold $\tau_c$, area threshold $\tau_a$,
balanced parameter $\alpha$, number of iterations $T$.
\ENSURE Weak augmentation $A_w$, strong augmentation $A_s$.\\
\FOR{$t=1$ to $T$}
\STATE $\{(x_n, y_n)\}_{n=1}^{B} \gets X$, $\{u_m\}_{m=1}^{B} \gets U$
\STATE Pseudo label $q_m \gets$ \texttt{argmax} $f(A_w(u_m))$
\STATE \textbf{\{Update training status\}}
\STATE Update CBMB according to $(x_n,y_n)$, $(u_m,q_m)$
\STATE Update class-wise data quantity $E_c$ 
via Eq. \ref{equ:2}, \ref{equ:3}
\STATE Update class-wise entropy $e_c$ 
via Eq. \ref{equ:5}, \ref{equ:6}
\STATE Update sampling probability $\hat{s}$, $\hat{s}^u$ via Eq. \ref{equ:alpha}
\STATE Update sample-wise entropy $e_m$, entropy masks $M_{h}$, $M_{l}$ and  entropy selection threshold $\tau_e$  via Eq. \ref{equ:8}, \ref{equ:9}, \ref{equ:10}\\
\STATE \textbf{\{Sampling\}}
\STATE $\{(x_m^{s}, y_m^{s})\}_{m=1}^{B},\{u_m^{s}\}_{m=1}^{B} \gets$ Sample labeled and unlabeled data from CBMB following $\hat{s}$\\
\STATE \textbf{\{Selection and CamMix\}}
\STATE $\{u_m^\prime\}_{m=1}^{B}$, $\lambda  \gets$ \texttt{CamMix}($A_{s}(u_m)$,  $A_{w}(x_m^{s})$, $y_m^{s}$, $A_{w}(u_m^{s})$, $f$) get mixed data and loss weight  following $M_{h}$, $M_{l}$, $\tau_c$ and $\tau_a$
\STATE \textbf{\{Compute losses\}}
\STATE Generate the mask of pseudo label $M_u$
\STATE $L_{s} \gets \sum_{n=1}^{B} \mathcal{H}(f(A_w(x_n)), y_n)$
\STATE $L_{u}^{h} \gets \hat{s}^u \sum_{m=1}^{B} M_u(u_m) M_{h}(u_m) \mathcal{H}(f(u_m^\prime), q_m)$
\STATE $L_{u}^{l} \gets \hat{s}^u \sum_{m=1}^{B} M_u(u_m) M_{l}(u_m) \mathcal{H}(f(u_m^\prime), q_m)$
\STATE $L_{u^s}^{h} \gets \sum_{m=1}^{B} M_{h}(u_m) \mathcal{H}(f(u_m^\prime), y_m^{s})$
\STATE $L_{u^s}^{l} \gets \hat{s}^u\sum_{m=1}^{B} M_u(u_m^{s}) M_{l}(u_m) \mathcal{H}(f(u_m^\prime), q_m^s)$
\STATE $L = L_{s} + \lambda( L_{u}^{h}+L_{u}^{l})+ (1-\lambda)(L_{u^s}^{h}+L_{u^s}^{l})$
\STATE Update $f$ based on $\nabla L$ using SGD
\ENDFOR
\RETURN
\end{algorithmic}
\end{algorithm}

\renewcommand{\algorithmicrequire}{\textbf{Input:}}
\renewcommand{\algorithmicensure}{\textbf{Output:}}
\begin{algorithm}[h]
\caption{CamMix}
\label{alg:cammix}
\begin{algorithmic}[1]
\REQUIRE  Strong augmentation of unlabeled data $A_{s}(u_m)$, weak augmentation of sampled labeled data $A_{w}(x_m^{s})$, the label of sampled labeled data $y_m^{s}$, weak augmentation of sampled unlabeled data $A_{w}(u_m^{s})$, model $f$, high entropy mask $M_{h}$, low entropy mask $M_{l}$, CAM threshold $\tau_c$, area threshold $\tau_a$, functions in skimage \texttt{label}($\cdot$) and \texttt{regionprops}($\cdot$), the function of CutMix \texttt{Mix}($\cdot$).
\ENSURE Mixed data $\{u_m^\prime\}_{m=1}^{B}$, loss weight $\lambda$.\\
\FOR{$m=1$ to $B$}
\STATE $q_m^s \gets$ \texttt{argmax} $f(A_w(u_m^s))$ 
\STATE $CAM_m^u \gets$  \texttt{GradCAM}$(A_{w}(u_m^{s}), q_m^s)$
\STATE $S_m^u \gets $ \texttt{int}$(CAM_m^u > \tau_c)$
\STATE $P_m^u \gets$ \texttt{max}(\texttt{regionprops}(\texttt{label}($S_m^u$))) get largest connected region
\IF{the area ratio of $P_m^u < \tau_a$}
\STATE $bbox_m^u \gets$ Random crop of $A_{w}(u_m^{s})$
\ELSE
\STATE $bbox_m^u \gets$ The bounding box of $P_m^u$
\ENDIF
\STATE $bbox_m^x \gets$ Calculate the bounding box for $(A_{w}(x_m^{s}), y_m^s)$ using a similar method in steps 3-10.
\STATE $u_m^\prime \gets$ \texttt{Mix}($A_{s}(u_m)$, $A_{w}(x_m^{s})$ or $A_{w}(u_m^{s})$) following $bbox_m^x$, $M_{h}(u_m)$, $bbox_m^u$, $M_{l}(u_m)$\\
\STATE $(1-\lambda_m) \gets$ The area ratio of  $bbox_m^x$ or $bbox_m^u$
\ENDFOR
\STATE $\lambda \gets$ The average of $\lambda_m$
\RETURN Mixed data $\{u_m^\prime\}_{m=1}^{B}$, loss weight $\lambda$
\end{algorithmic}
\end{algorithm}
\begin{itemize}
\item \textbf{CIFAR10/100-LT} CIFAR-10/100~\cite{krizhevsky2009learning} are originally class-balanced datasets, each containing 500/5000 samples across 10 and 100 classes respectively. All images are $32\times32$ in size. Following previous work~\cite{oh2022daso}, we sample the training data to create imbalanced versions of the datasets. We employ different sampling ratios for labeled and unlabeled data to achieve various data distributions, including $\gamma_l=\gamma_u$ and $\gamma_l \neq \gamma_u$ scenarios. The test set contains $10k$ samples with a balanced class distribution. The CIFAR dataset can be downloaded from \href{https://www.cs.toronto.edu/~kriz/cifar.html}{https://www.cs.toronto.edu/~kriz/cifar.html}.
\item \textbf{STL10-LT} The STL-10~\cite{coates2011analysis} dataset consists of 5000 class-balanced labeled data and 1000k unlabeled data with an unknown distribution. To make an imbalanced version of the dataset, we only sample the labeled data, while the distribution of unlabeled data naturally differs from that of labeled data, i.e.,$\gamma_l \neq \gamma_u$. All images are $96\times96$ in size and the dataset can be downloaded from \href{https://cs.stanford.edu/~acoates/stl10/}{https://cs.stanford.edu/~acoates/stl10/}.
\item \textbf{ImageNet-127}  ImageNet-127~\cite{fan2022cossl} is naturally an imbalanced dataset, thus it doesn't require any further processing. Moreover, it has an imbalanced test set, which can validate scenario \emph{3)}. To conserve computation resources, all images are down-sampled to $32\times32$ or $64\times64$ in size and the dataset can be downloaded from \href{https://image-net.org/download-images}{https://image-net.org/download-images}.
\end{itemize}

\noindent\textbf{Implementation details.} 
Following previous training protocol~\cite{oh2022daso}, we conduct our experiments  on CIFAR10-LT, CIFAR100-LT and STL10-LT using Wide ResNet-28-2~\cite{zagoruyko2016wide}, and  on ImageNet-127 using ResNet-50~\cite{he2016deep}. We train the model with a batch size of 64 for $250k$ iterations, with an evaluation every 500 iterations. We use SGD with momentum as our optimizer and adopt a cosine learning rate decay strategy by setting the learning rate to $ \eta cos(\frac{7\pi t}{16T})$, where $\eta$ is the initial learning rate, $t$ is the current iteration number and $T$ is the total number of iterations. We set the balance parameter $\alpha=0.5$ on CIFAR10-LT, CIFAR100-LT and STL10-LT, and set it to 0.2 on ImageNet-127. We set all EMA update weights as $\lambda = \lambda_d = \lambda_e = \lambda_\tau = 0.999$. The CAM threshold $\tau_{c}$ and area threshold $\tau_{a}$ are set to 0.8 and 0.1, respectively. The epoch number for starting to estimate the data quantity and entropy of unlabeled data is set to 5. We designate the final block as the CAM layer. We adopt Softmax$(\cdot)$ as the mapping function $\delta(\cdot)$.
Our experiments are conducted on one NVIDIA Tesla V100 with the CentOS 7 system, using PyTorch 1.11.0 and Torchvision 0.12.0.

\section{Pseudo-code for Our BEM Algorithm}
We define the pseudo-code for our BEM and CamMix algorithm in Alg.~\ref{alg:2_BEM} and ~\ref{alg:cammix}, respectively.
\label{sec:2}


\begin{table*}[!h]
\setlength\tabcolsep{2pt}
\begin{center}
\caption{Comparison of test accuracy with combinations of different baseline models under $\gamma_l \neq \gamma_u$ setup on CIFAR10-LT and STL10-LT. The $\gamma_l$ is fixed to 100 for CIFAR10-LT, and the $\gamma_l$ is set to 10 and 20 for STL10-LT. The best results for each diversion are in \textbf{bold}.}
\label{tab:sota3}
\resizebox{0.95\textwidth}{!}{
\begin{tabular}{llcccccccc}
\toprule[1pt]
& & \multicolumn{4}{c}{CIFAR10-LT($\gamma_l \neq \gamma_u$)}& \multicolumn{4}{c}{STL10-LT($\gamma_u = N/A$)}\\
& & \multicolumn{2}{c}{$\gamma_u=1$(uniform)}&\multicolumn{2}{c}{$\gamma_u=1/100$(reversed)}&\multicolumn{2}{c}{$\gamma_l=10$}& \multicolumn{2}{c}{$\gamma_l=20$}\\
\cmidrule(r){3-4} \cmidrule(r){5-6} \cmidrule(r){7-8} \cmidrule(r){9-10}
& &$N_1=500$ &$N_1=1500$ &$N_1=500$ &$N_1=1500$ &$N_1=150$ &$N_1=450$ &$N_1=150$ &$N_1=450$ \\
Algorithm& &$M_1=4000$ &$M_1=3000$ &$M_1=4000$ &$M_1=3000$ &$M=100k$ &$M=100k$ &$M=100k$ &$M=100k$ \\
\cmidrule(r){1-2} \cmidrule(r){3-4} \cmidrule(r){5-6} \cmidrule(r){7-8} \cmidrule(r){9-10}
\multicolumn{2}{l}{FixMatch~\cite{sohn2020fixmatch}}&73.0\footnotesize{±3.81}& 81.5\footnotesize{±1.15}& 62.5\footnotesize{±0.94}& 71.8\footnotesize{±1.70}& 56.1\footnotesize{±2.32}& 72.4\footnotesize{±0.71}& 47.6\footnotesize{±4.87}& 64.0\footnotesize{±2.27}\\
\multicolumn{2}{l}{\quad w/DASO~\cite{oh2022daso}}&86.6\footnotesize{±0.84}& 88.8\footnotesize{±0.59} &\textbf{71.0 }\footnotesize{±0.95} &\textbf{80.3}\footnotesize{±0.65}& \textbf{70.0}\footnotesize{±1.19}&78.4\footnotesize{±0.80}& \textbf{65.7}\footnotesize{±1.78} &75.3\footnotesize{±0.44}\\
\multicolumn{2}{l}{\quad w/BEM(Ours) }&\textbf{86.8}\footnotesize{±0.47}&\textbf{89.1}\footnotesize{±0.75}&70.0\footnotesize{±1.72}&79.1\footnotesize{±0.77}&
68.3\footnotesize{±1.15}&\textbf{81.2}\footnotesize{±1.42}&61.6\footnotesize{±0.98} &\textbf{76.0}\footnotesize{±1.51}\\
\cmidrule(r){1-2} \cmidrule(r){3-4} \cmidrule(r){5-6} \cmidrule(r){7-8} \cmidrule(r){9-10}
\multicolumn{2}{l}{\quad w/LA~\cite{menon2020long}+DASO~\cite{oh2022daso}}&84.6\footnotesize{±2.04}&86.8\footnotesize{±0.76} &\textbf{72.6 }\footnotesize{±0.38} &78.5\footnotesize{±1.31}& 72.7\footnotesize{±1.45}&79.7\footnotesize{±0.44}& \textbf{66.8}\footnotesize{±0.62} &75.7\footnotesize{±0.50}\\
\multicolumn{2}{l}{\quad w/LA~\cite{menon2020long}+BEM(Ours) }&\textbf{85.3}\footnotesize{±0.31}&\textbf{88.5}\footnotesize{±0.65}&70.9\footnotesize{±1.69}&\textbf{79.8}\footnotesize{±1.37}&
\textbf{72.9}\footnotesize{±0.38}&\textbf{81.8}\footnotesize{±0.76}&65.7\footnotesize{±0.25} &\textbf{76.8}\footnotesize{±1.87}\\
\cmidrule(r){1-2} \cmidrule(r){3-4} \cmidrule(r){5-6} \cmidrule(r){7-8} \cmidrule(r){9-10}
\multicolumn{2}{l}{\quad w/ABC~\cite{lee2021abc}+DASO~\cite{oh2022daso}}&85.2\footnotesize{±1.56}& 88.4\footnotesize{±0.82} &70.1\footnotesize{±1.25} &79.8\footnotesize{±0.21}& 71.8\footnotesize{±1.17}&78.4\footnotesize{±0.58}& \textbf{67.3}\footnotesize{±2.06} &75.9\footnotesize{±0.43}\\
\multicolumn{2}{l}{\quad w/ABC~\cite{lee2021abc}+BEM(Ours) }&\textbf{85.9}\footnotesize{±0.33}&\textbf{89.0}\footnotesize{±0.67}&\textbf{71.2}\footnotesize{±0.58}&\textbf{80.1}\footnotesize{±0.96}&
\textbf{73.1}\footnotesize{±1.68}&\textbf{81.4}\footnotesize{±1.29}&66.4\footnotesize{±1.93} &\textbf{76.7}\footnotesize{±1.12}\\
\cmidrule(r){1-2} \cmidrule(r){3-4} \cmidrule(r){5-6} \cmidrule(r){7-8} \cmidrule(r){9-10}
\multicolumn{2}{l}{\quad w/ACR~\cite{wei2023towards}}&92.1\footnotesize{±0.18}&93.5\footnotesize{±0.11}&85.0\footnotesize{±0.09}&89.5\footnotesize{±0.17}&77.1\footnotesize{±0.24}&83.0\footnotesize{±0.32}&75.1\footnotesize{±0.70}&81.5\footnotesize{±0.25}\\
\multicolumn{2}{l}{\quad w/ACR~\cite{wei2023towards}+w/BEM(Ours) }&\textbf{94.3}\footnotesize{±0.14}&\textbf{95.1}\footnotesize{±0.56}&\textbf{85.5}\footnotesize{±0.21}&\textbf{89.8}\footnotesize{±0.12}&\textbf{79.3}\footnotesize{±0.34}&\textbf{84.2}\footnotesize{±0.56}&\textbf{75.9}\footnotesize{±0.15}&\textbf{82.3}\footnotesize{±0.23}\\
\toprule[1pt]
\end{tabular}
}
\end{center}
\end{table*}

\begin{table}[!h]
\setlength\tabcolsep{2pt}
\begin{center}
\caption{Comparison of test accuracy with combinations of different baseline models under $\gamma_l \neq \gamma_u$ setup on CIFAR100-LT. The $\gamma_l$ is fixed to 10. The best results for each diversion are in \textbf{bold}.}
\label{tab:CIFAR100}
\resizebox{0.48\textwidth}{!}{
\begin{tabular}{llcccc}
\toprule[1pt]
& & \multicolumn{4}{c}{CIFAR100-LT($\gamma_l \neq \gamma_u$)}\\
& & \multicolumn{2}{c}{$\gamma_u=1$(uniform)}&\multicolumn{2}{c}{$\gamma_u=1/10$(reversed)}\\
\cmidrule(r){3-4} \cmidrule(r){5-6}
& &$N_1=50$ &$N_1=150$ &$N_1=50$ &$N_1=150$  \\
Algorithm& &$M_1=400$ &$M_1=300$ &$M_1=400$ &$M_1=300$\\
\cmidrule(r){1-2} \cmidrule(r){3-4} \cmidrule(r){5-6}
\multicolumn{2}{l}{FixMatch~\cite{sohn2020fixmatch}}&45.5\footnotesize{±0.71}& 58.1\footnotesize{±0.72}& 44.2\footnotesize{±0.43}& 57.3\footnotesize{±0.19}\\
\multicolumn{2}{l}{\quad w/DASO~\cite{oh2022daso}}&53.9\footnotesize{±0.66}& 61.8\footnotesize{±0.98}& \textbf{51.0}\footnotesize{±0.19}& 60.0\footnotesize{±0.31}\\
\multicolumn{2}{l}{\quad w/BEM(Ours)}&\textbf{54.8}\footnotesize{±0.55}& \textbf{63.6}\footnotesize{±0.91}& 50.8\footnotesize{±0.25}& \textbf{60.7}\footnotesize{±0.12}\\
\cmidrule(r){1-2} \cmidrule(r){3-4} \cmidrule(r){5-6} 
\multicolumn{2}{l}{\quad w/LA~\cite{menon2020long}+DASO~\cite{oh2022daso}}&54.7\footnotesize{±0.40}& 62.4\footnotesize{±1.06}& 51.1\footnotesize{±0.12}& 60.5\footnotesize{±0.23}\\
\multicolumn{2}{l}{\quad w/LA~\cite{menon2020long}+BEM(Ours)}&\textbf{56.5}\footnotesize{±0.43}& \textbf{64.1}\footnotesize{±0.87}& \textbf{51.7}\footnotesize{±0.20}& \textbf{61.3}\footnotesize{±0.17}\\
\cmidrule(r){1-2} \cmidrule(r){3-4} \cmidrule(r){5-6} 
\multicolumn{2}{l}{\quad w/ABC~\cite{lee2021abc}+DASO~\cite{oh2022daso}}&53.4\footnotesize{±0.53}& 62.4\footnotesize{±0.61}& \textbf{51.2}\footnotesize{±0.19}& 60.8\footnotesize{±0.39}\\
\multicolumn{2}{l}{\quad w/ABC~\cite{lee2021abc}+BEM(Ours)}&\textbf{55.2}\footnotesize{±0.35}& \textbf{64.7}\footnotesize{±0.87}& 51.1\footnotesize{±0.10}& \textbf{61.4}\footnotesize{±0.29}\\
\cmidrule(r){1-2} \cmidrule(r){3-4} \cmidrule(r){5-6} 
\multicolumn{2}{l}{\quad w/ACR~\cite{wei2023towards}}&66.0\footnotesize{±0.25}&73.4\footnotesize{±0.22}&57.0\footnotesize{±0.46}&67.6\footnotesize{±0.12}\\
\multicolumn{2}{l}{\quad w/ACR~\cite{wei2023towards}+BEM(Ours) }&\textbf{68.1}\footnotesize{±0.34}&\textbf{75.9}\footnotesize{±0.49}&\textbf{58.0\footnotesize{±0.28}}&\textbf{68.4\footnotesize{±0.13}}\\

\toprule[1pt]
\end{tabular}
}
\end{center}
\vspace{-10pt}
\end{table}

\begin{table}[!t]
\setlength\tabcolsep{2pt}
\begin{center}
\caption{Fine-grained results on CIFAR10-LT with $N_1=1500,M_1=3000,\gamma_l=100$.}
\label{tab:fine-grained}
\resizebox{0.48\textwidth}{!}{
\begin{tabular}{lcccccccccccc}
\toprule[1pt]
& \multicolumn{4}{c}{Consistent($\gamma_u=100$)}& \multicolumn{4}{c}{Uniform($\gamma_u=1$)}& \multicolumn{4}{c}{Reversed($\gamma_u=1/100$)}\\
\cmidrule(r){2-5} \cmidrule(r){6-9} \cmidrule(r){10-13}
Algorithm& Many & Medium & Few & All & Many & Medium & Few & All & Many & Medium & Few & All \\
\cmidrule(r){1-1}  \cmidrule(r){2-5} \cmidrule(r){6-9} \cmidrule(r){10-13}
DASO & \textbf{95.1} & 78.6  &60.4 &78.1 &89.6 &84.4 &85.7 &86.3 &84.0 &71.6 &68.2 &74.3\\
BEM & 94.7 & 78.0  &67.0 &79.8 &91.7 &88.1 &90.7 &89.4 &82.3 &80.2 &73.3 &78.7\\
ACR & 93.9 & 81.6  &75.3 &83.4 &92.8 &90.6 &97.9 &93.5 &90.7 &83.8 &\textbf{96.4} &89.7\\
ACR+BEM & 92.3 & \textbf{83.3}  &\textbf{81.9} &\textbf{85.4} &\textbf{95.4} &\textbf{93.1} &\textbf{98.0} &\textbf{95.3} &\textbf{90.9} &\textbf{84.9} &95.8 &\textbf{89.9}\\
\toprule[1pt]
\end{tabular}
}
\end{center}
\vspace{-20pt}
\end{table}

\begin{table}[!t]
\begin{center}
\caption{Ablation study on different sampling strategies. EFF. denotes the effective number. }
\label{tab:sampling}
\resizebox{0.48\textwidth}{!}{
\begin{tabular}{lccc|cc}
\toprule[1pt]
~&CBMB &ESS  &EFF. &\multicolumn{1}{c}{C10}&  \multicolumn{1}{c}{STL10}\\
\hline
Random& & & & 72.1 & 65.0 \\
Quantity-based&\checkmark & &\checkmark & 74.9 & 66.5 \\
Entropy-based& &\checkmark & & 74.4 & 65.9 \\
w/o effective number&\checkmark &\checkmark & & 75.2 & 67.3 \\
Ours&\checkmark &\checkmark &\checkmark & \textbf{75.7} & \textbf{68.3} \\
\toprule[1pt]
\end{tabular}
}
\end{center}
\vspace{-10pt}
\end{table}

\begin{table}[!t]
\begin{center}
\caption{Ablation study on updating strategies of entropy selection threshold $\tau_e$.}
\label{tab:tau_e_updating}
\begin{tabular}{lcc}
\toprule[1pt]
& \multicolumn{1}{c}{C10}&  \multicolumn{1}{c}{STL10}\\
\hline
\multicolumn{1}{l}{Baseline}&  67.8 & 56.1 \\
\multicolumn{1}{l}{$\tau_e=0.1$}&  74.7 & 66.6 \\
\multicolumn{1}{l}{$\tau_e=0.2$}&  75.2 & 67.2 \\
\multicolumn{1}{l}{$\tau_e=0.4$}&  75.1 & 66.9 \\
\multicolumn{1}{l}{$\tau_e=0.6$}&  74.4 & 66.4 \\
\multicolumn{1}{l}{w/ ours}&  \textbf{75.7} & \textbf{68.3} \\
\toprule[1pt]
\end{tabular}
\end{center}
\vspace{-20pt}
\end{table}

\section{Additional Experiment Results}
\label{sec:3_res}
In this section,  we conduct a series of additional experiments to further demonstrate the effectiveness of our BEM.

\noindent\textbf{More results  with re-balancing methods when $\gamma_l \neq \gamma_u$.} 
We present the results of combining with FixMatch and ACR under  $\gamma_l \neq \gamma_u$ setup in Tab.~\ref{tab:sota2}. As shown in Tab.~\ref{tab:sota3}, we further combine our BEM with more re-balancing methods, including LA and ABC.  
Without incorporating any re-balancing method, BEM's performance is weaker than DASO in some settings, particularly in the reversed setting. After combining two re-balancing methods, BEM outperforms DASO in almost all settings. Further integration with ACR achieves the state-of-the-art results in all scenarios with an average 31.5\% performance gain. In summary, our method needs to combine with re-balancing methods to enhance the re-balancing ability in challenging scenarios, and it in turn complements these methods.


\noindent\textbf{More results on CIFAR100-LT.} 
We also conduct experiments on CIFAR100-LT under $\gamma_l \neq \gamma_u$ setup in Tab.~\ref{tab:CIFAR100}. 
Results show that our BEM outperforms DASO in almost all settings. By integrating with ACR, we can achieve the best results in all scenarios (32.7\% accuracy gain). It further demonstrates that the complementation of BEM  can boost the performance of most re-balancing methods.

\begin{table*}[!t]
\begin{center}
\caption{Comparison of test accuracy on balanced datasets with combinations of different SSL methods, including MeanTeacher, FixMatch, FlexMatch and SoftMatch.}
\label{tab:balanced dataset}
\resizebox{0.95\textwidth}{!}{
\begin{tabular}{llcccccccc}
\toprule[1pt]
& & \multicolumn{3}{c}{CIFAR-10}&  \multicolumn{3}{c}{CIFAR-100}&  \multicolumn{2}{c}{STL-10}\\
\cmidrule(r){3-5} \cmidrule(r){6-8}  \cmidrule(r){9-10} 
Algorithm& & 40 & 250 & 4000 & 400 & 2500 & 10000 & 40 & 1000 \\
\cmidrule(r){1-2} \cmidrule(r){3-5} \cmidrule(r){6-8}  \cmidrule(r){9-10} 
\multicolumn{2}{l}{MeanTeacher\cite{tarvainen2017mean}}&  29.81\footnotesize{±1.60} & 62.54\footnotesize{±3.30} & 91.90\footnotesize{±0.21} & 18.89\footnotesize{±1.44}& 54.83\footnotesize{±1.06}& 68.25\footnotesize{±0.23}& 28.28\footnotesize{±1.45}& 66.10\footnotesize{±1.37} \\
\multicolumn{2}{l}{\quad w/BEM(Ours)}&  \textbf{43.13}\footnotesize{±2.55} & \textbf{74.31}\footnotesize{±1.79} & \textbf{92.65}\footnotesize{±0.23} & \textbf{30.92}\footnotesize{±3.69}& \textbf{60.73}\footnotesize{±2.14}& \textbf{72.54}\footnotesize{±0.19}& \textbf{37.31}\footnotesize{±2.59}& \textbf{78.74}\footnotesize{±1.38} \\
\cmidrule(r){1-2} \cmidrule(r){3-5} \cmidrule(r){6-8}  \cmidrule(r){9-10} 
\multicolumn{2}{l}{FixMatch~\cite{sohn2020fixmatch}}&  92.53\footnotesize{±0.28}& 95.14\footnotesize{±0.05} & 95.79\footnotesize{±0.08} & 53.58\footnotesize{±0.82}&  72.97\footnotesize{±0.16}& 77.80\footnotesize{±0.12} & 64.03\footnotesize{±4.14} & 93.75\footnotesize{±0.33}\\
\multicolumn{2}{l}{\quad w/BEM(Ours)}& \textbf{93.96}\footnotesize{±0.37} & \textbf{95.37}\footnotesize{±0.03} & \textbf{95.93}\footnotesize{±0.11} & \textbf{55.24}\footnotesize{±0.93}& \textbf{73.12}\footnotesize{±0.14} & \textbf{77.95}\footnotesize{±0.11} & \textbf{66.45}\footnotesize{±3.29} & \textbf{93.98}\footnotesize{±0.65}  \\
\cmidrule(r){1-2} \cmidrule(r){3-5} \cmidrule(r){6-8}  \cmidrule(r){9-10} 
\multicolumn{2}{l}{FlexMatch~\cite{zhang2021flexmatch}}&  95.03\footnotesize{±0.06}& 95.03\footnotesize{±0.09} & 95.81\footnotesize{±0.01} & 60.06\footnotesize{±1.62}& 73.51\footnotesize{±0.20}& 78.10\footnotesize{±0.09} & 70.85\footnotesize{±0.01} & 94.23\footnotesize{±1.62} \\
\multicolumn{2}{l}{\quad w/BEM(Ours)}& \textbf{95.08}\footnotesize{±0.09} & \textbf{95.21}\footnotesize{±0.04} & \textbf{95.98}\footnotesize{±0.01} & \textbf{60.83}\footnotesize{±0.98} & \textbf{73.94}\footnotesize{±0.18}& \textbf{78.72}\footnotesize{±0.11}& \textbf{72.11}\footnotesize{±0.03}& \textbf{94.39}\footnotesize{±1.54}\\
\cmidrule(r){1-2} \cmidrule(r){3-5} \cmidrule(r){6-8}  \cmidrule(r){9-10} 
\multicolumn{2}{l}{SoftMatch~\cite{chen2023softmatch}}&  95.09\footnotesize{±0.12} & 95.18\footnotesize{±0.09} & 95.96\footnotesize{±0.02}& 62.90\footnotesize{±0.77}&  73.34\footnotesize{±0.25} & 77.97\footnotesize{±0.03} & 78.58\footnotesize{±3.48}& 94.27\footnotesize{±0.24} \\
\multicolumn{2}{l}{\quad w/BEM(Ours)}&  \textbf{95.11}\footnotesize{±0.08} & \textbf{95.37}\footnotesize{±0.06} & \textbf{96.12}\footnotesize{±0.07} & \textbf{63.13}\footnotesize{±0.92}& \textbf{73.56}\footnotesize{±0.08}& \textbf{78.14}\footnotesize{±0.08}& \textbf{79.09}\footnotesize{±3.87}& \textbf{94.43}\footnotesize{±0.38} \\
\toprule[1pt]
\end{tabular}
}
\end{center}
\end{table*}

\noindent\textbf{Fine-grained results.}
In this experiment, we present the fine-grained results in Tab.~\ref{tab:fine-grained}. We compare our BEM with DASO and ACR in three settings. Our method surpasses DASO in all scenarios and further enhances the state-of-the-art method (ACR). In particular, our method significantly improves the performance of few-shot classes at the cost of negligible drop on head classes in the consistent setting. Moreover, in all settings, our method shows a large improvement in medium classes, which is brought by entropy-based learning.

\noindent\textbf{BEM on balanced datasets.}
To verify the effect of our BEM on balanced datasets, we conduct experiments on balanced datasets with combinations of different SSL methods, including MeanTeacher, FixMatch, FlexMatch and SoftMatch.
Specifically, we set $\alpha=0$, meaning that we only consider the differences in class-wise uncertainty distribution. As shown in Tab.~\ref{tab:balanced dataset}, our BEM enhances the performance of all baseline models, particularly the MeanTeacher, where our method gains an average of 21.4\%, 26.9\% and 25.0\% improvement for three datasets. This demonstrates the potential of class-wise uncertainty re-balancing in enhancing model performance for balanced datasets.




\noindent\textbf{Ablation study on sampling strategies.}
To evaluate the effect of our sampling strategy, we conduct a series of experiments by replacing the sampling function. Results are summarized in Tab.~\ref{tab:sampling}. Random sampling only improves performance slightly. Then, we split the class-balanced entropy-based sampling function and find that the results drop on both datasets. Further, we replace the effective number with the common number. Results indicate the effective number more accurately measures the class distribution of datasets.


\noindent\textbf{Ablation on the updating strategy of entropy threshold $\tau_e$.}
As shown in Tab.~\ref{tab:tau_e_updating}, we perform experiments to validate the effect of the entropy threshold $\tau_e$ updating strategy. When we filter the entropy mask with fixed thresholds, the performance decreases and becomes unstable. Our EMA updating strategy achieves the best result, indicating that it adaptively adjusts the threshold following the training status of the model.

\noindent\textbf{Ablation study on parameter $\alpha$.}
As shown in  Tab.~\ref{tab:alpha}, we verify the effect of $\alpha$ to balance the effective number and entropy in Eq.~\ref{equ:alpha}. Results show the best $\alpha$ on CIFAR10-LT and STL10-LT are 0.5 and 0.3, respectively.  The visualization of sampling rate and class accuracy can be seen in Appendix~\textcolor{red}{E}.

\begin{table}[!t]
\begin{center}
\caption{Ablation study on  $\alpha$.}
\begin{tabular}{lcc}
\toprule[1pt]
& \multicolumn{1}{c}{C10}&  \multicolumn{1}{c}{STL10}\\
\hline
\multicolumn{1}{l}{1.0}&74.7&67.0  \\
\multicolumn{1}{l}{0.7}&75.5&67.3  \\
\multicolumn{1}{l}{0.5}& \textbf{75.7} & 68.3  \\
\multicolumn{1}{l}{0.3}& 74.4&\textbf{68.5}  \\
\multicolumn{1}{l}{0}&73.8&67.5   \\
\toprule[1pt]
\end{tabular}
\label{tab:alpha}
\end{center}
\vspace{-20pt}
\end{table}

\noindent\textbf{Ablation study on CAM threshold $\tau_c$.} 
In Tab.~\ref{tab:CAM}, we study the effect of CAM threshold $\tau_c$ on selected region. Results show that 0.8 is the best threshold on both datasets. It indicates that the precise selection of relevant regions is more advantageous for re-balancing long-tailed datasets. 

\begin{table}[!t]
\begin{center}
\caption{Ablation study on $\tau_{c}$.}
\begin{tabular}{lcc}
\toprule[1pt]
& \multicolumn{1}{c}{C10}&  \multicolumn{1}{c}{STL10}\\
\hline
\multicolumn{1}{l}{0.9}&  73.0 & 65.8 \\
\multicolumn{1}{l}{0.8}&  \textbf{75.7} & \textbf{68.3} \\
\multicolumn{1}{l}{0.6}&  74.4 & 67.3 \\
\multicolumn{1}{l}{0.4}&  71.5 & 64.6 \\
\multicolumn{1}{l}{0.2}&  69.3 & 61.3 \\
\toprule[1pt]
\end{tabular}
\label{tab:CAM}
\end{center}
\vspace{-20pt}
\end{table}

\noindent\textbf{Ablation study on the adding weight $\beta$.}
We conduct experiments to test the impact of the adding weight $\beta$ in the equation $e_c = \beta e_c^u + (1-\beta)e_c^x$. The results in Tab.~\ref{tab:beta} indicate that weight addition has minimal impact. So we remove this parameter to simplify the number of hyperparameters.

\begin{table}[!t]
\begin{center}
\caption{Ablation study on the adding weight $\beta$.}
\label{tab:beta}
\begin{tabular}{lcc}
\toprule[1pt]
\multicolumn{1}{c}{$\beta$}& \multicolumn{1}{c}{C10}&  \multicolumn{1}{c}{STL10}\\
\hline
\multicolumn{1}{l}{0.7}& 75.1 & 67.7 \\ 
\multicolumn{1}{l}{0.5}& 75.7 & \textbf{68.3}\\
\multicolumn{1}{l}{0.3}&  \textbf{75.8}& 68.1\\
\multicolumn{1}{l}{0.1}&  74.9& 67.9\\
\toprule[1pt]
\end{tabular}
\end{center}
\vspace{-20pt}
\end{table}

\noindent\textbf{More comparison with class-wise data mixing methods.}
We conduct additional experiments to compare our BEM with other class-wise data mixing methods~\cite{zhong2021improving, xu2021towards}. The results in Tab.~\ref{tab:datamix_sup} show that BEM outperforms them. We infer that these class-wise mixup methods are limited in not considering the uncertainty issue in LTSSL. 

\begin{table}[!t]
\begin{center}
\caption{More comparison with class-wise data mixing methods.}
\label{tab:datamix_sup}
\begin{tabular}{llcc}
\toprule[1pt]
& & \multicolumn{1}{c}{C10}&  \multicolumn{1}{c}{STL10}\\
\hline
\multicolumn{2}{l}{FixMatch}&  67.8 & 56.1 \\
\multicolumn{2}{l}{\quad w/UniMix~\cite{xu2021towards}}& 72.9 & 66.0 \\
\multicolumn{2}{l}{\quad w/MiSLAS~\cite{zhong2021improving}}& 73.4 & 66.2 \\
\multicolumn{2}{l}{\quad w/Ours}& \textbf{75.7} & \textbf{68.3}\\
\toprule[1pt]
\end{tabular}
\end{center}
\vspace{-20pt}
\end{table}

\begin{table}[!t]
\begin{center}
\caption{Comparison with AREA on supervised learning.}
\label{tab:AREA}
\begin{tabular}{lcccc}
\toprule[1pt]
& \multicolumn{2}{c}{C10-LT}&  \multicolumn{2}{c}{C100-LT}\\
\cmidrule(r){2-3} \cmidrule(r){4-5}
$\gamma$&200 &50& 200 & 50 \\
\hline
CE&  65.7 & 74.8&34.8 & 43.9 \\
AREA~\cite{chen2023area}& \textbf{75.0}& 82.7&\textbf{43.9} & \textbf{51.8} \\
Ours & 74.7 & \textbf{83.0} & 40.3 & 49.7  \\
\toprule[1pt]
\end{tabular}
\end{center}
\vspace{-20pt}
\end{table}

\noindent\textbf{Comparison with AREA.} We compare our BEM with AREA~\cite{chen2023area}, which is a fully supervised learning method in long-tailed learning. Our BEM is different from AREA in three aspects:
\textbf{1) Motivation:} 
AREA does not consider class-wise uncertainty. It optimizes the re-weighting strategy, which only focuses on data quantity, by exploring the spanned space of each class and relations between samples. While we propose to re-balance the class distribution of both data quantity and uncertainty, which is more suitable for LTSSL.
\textbf{2) Task:} 
AREA focuses only on the class imbalance issue in the supervised learning diagram. While our method is specifically designed for LTSSL to further address the issue of uncertainty in unlabeled sample predictions, which can not be achieved by AREA. We also apply our BEM to supervised learning. Tab.~\ref{tab:AREA} shows that BEM is competitive with AREA, demonstrating its flexibility and superiority.
\textbf{3)  Design:}  AREA is based on the re-weighting strategy, using the  effective area as class-wise weights in cross-entropy loss. While BEM is primarily based on re-sampling, where we use class-wise data quantity and uncertainty as sampling criteria for CamMix.


\section{Additional Visualization Analysis}
\label{sec:4_vis}
In this section, we provide additional visualization analysis to better understand our approach.

\noindent\textbf{Visualization of confusion matrices on test set.}
We compare the confusion matrices of the prediction from the test set. We conduct experiments on CIFAR10-LT in the consistent scenario and apply our BEM to FixMatch and ACR, respectively. As shown in Fig.~\ref{fig:confusion matrix}, the prediction of FixMatch is significantly biased towards the head classes, resulting in poor performance of the tail classes. Our method greatly alleviates this bias, improving both the tail performance and overall performance. ACR achieves good results in various classes, and our method further improves the performance of the tail classes, demonstrating the superiority and versatility of our method.

\begin{figure}[!t]
    \centering
    \includegraphics[width=\linewidth]{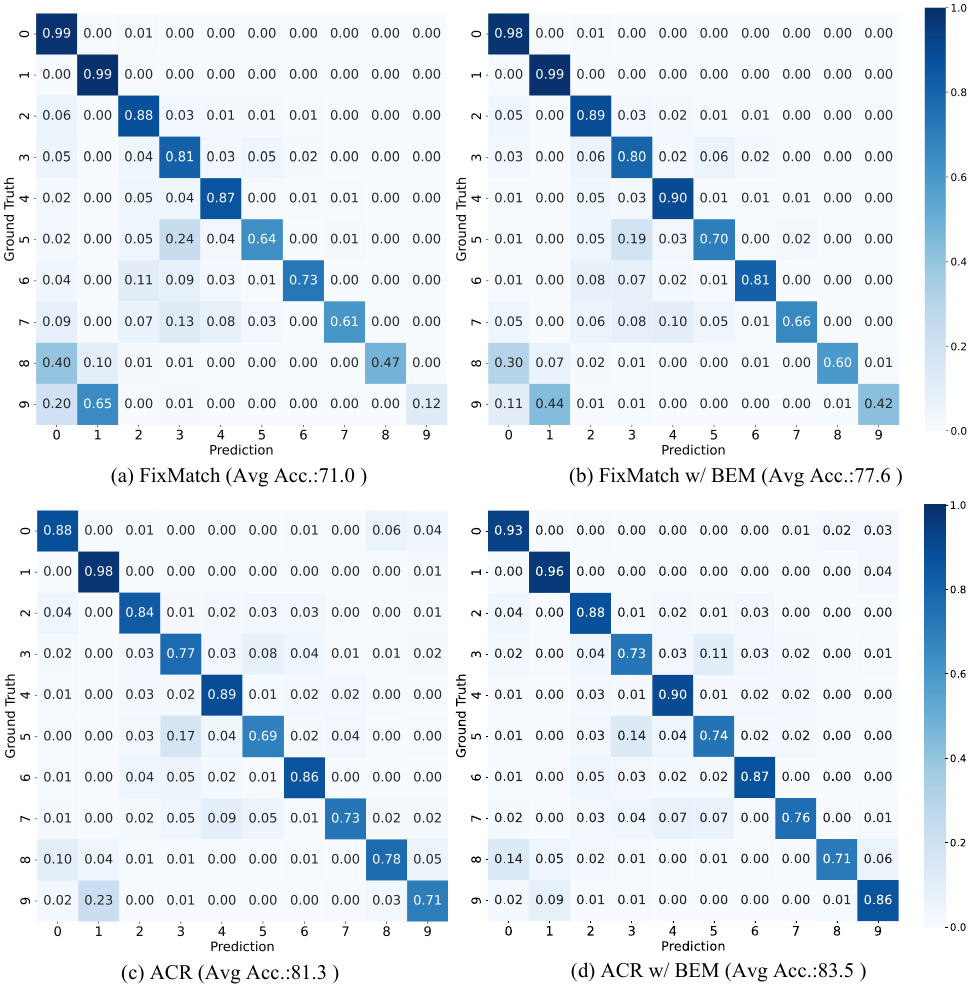}
    \caption{The confusion matrices of the test set on CIFAR10-LT under $\gamma_l=\gamma_u$ setup.}
    \label{fig:confusion matrix}
\end{figure}

\noindent\textbf{Visualization of precision and recall on the test set.}
We analyze the precision and recall on the test set to further verify the effect of our BEM. As shown in Fig.~\ref{fig:prerecall}, we apply our method to FixMatch and ACR.
The results show that the recall of tail classes achieves significant gains by combining our BEM with both models.
 
\begin{figure}[!t]
  \centering
  \begin{subfigure}[b]{0.23\textwidth}
\includegraphics[width=\textwidth]{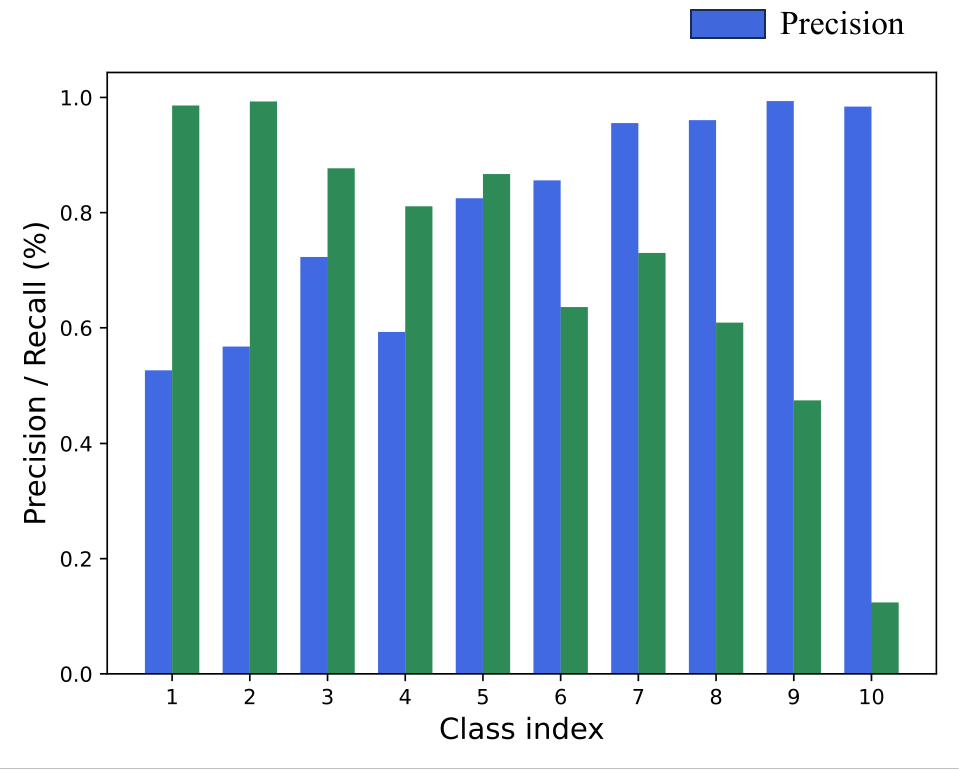}
    \caption{FixMatch}
    \label{fig:image1}
  \end{subfigure}
  \hfill
  \begin{subfigure}[b]{0.23\textwidth}
    \includegraphics[width=\textwidth]{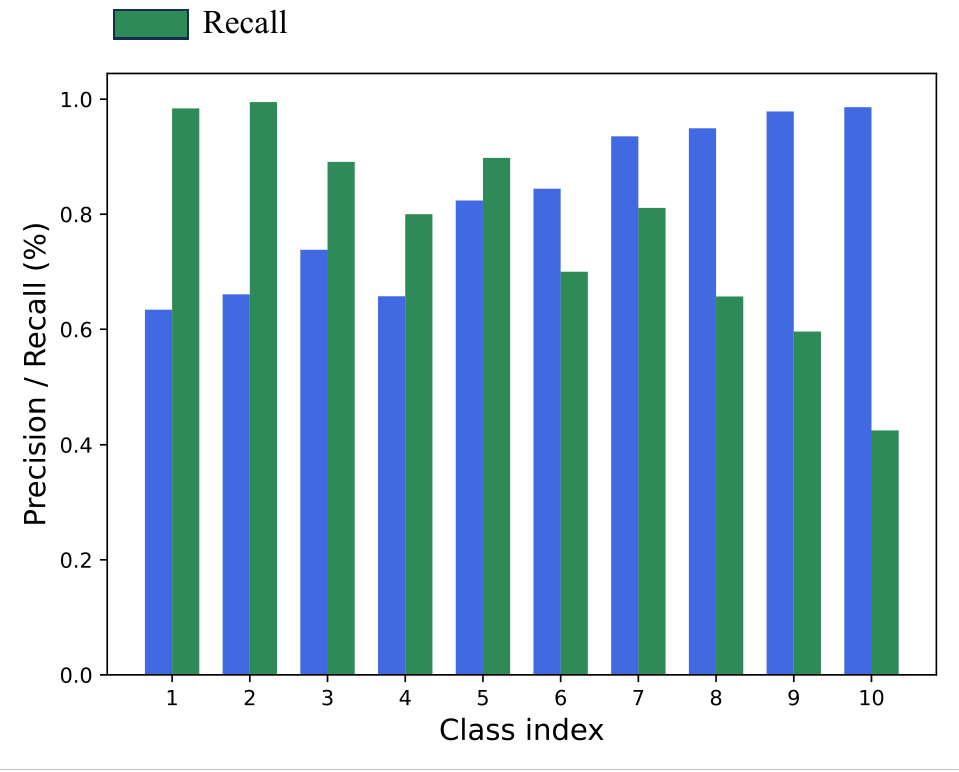}
    \caption{FixMatch w/BEM}
    \label{fig:image2}
  \end{subfigure}
  \\
  \begin{subfigure}[b]{0.23\textwidth}
    \includegraphics[width=\textwidth]{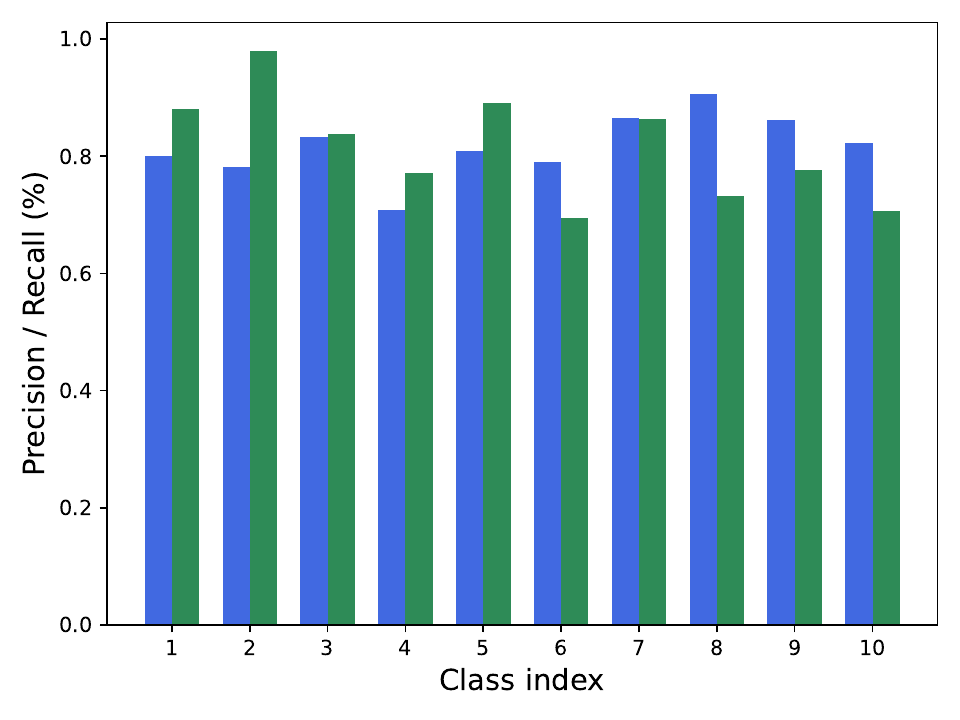}
    \caption{ACR}
    \label{fig:image3}
  \end{subfigure}
  \hfill
  \begin{subfigure}[b]{0.23\textwidth}
    \includegraphics[width=\textwidth]{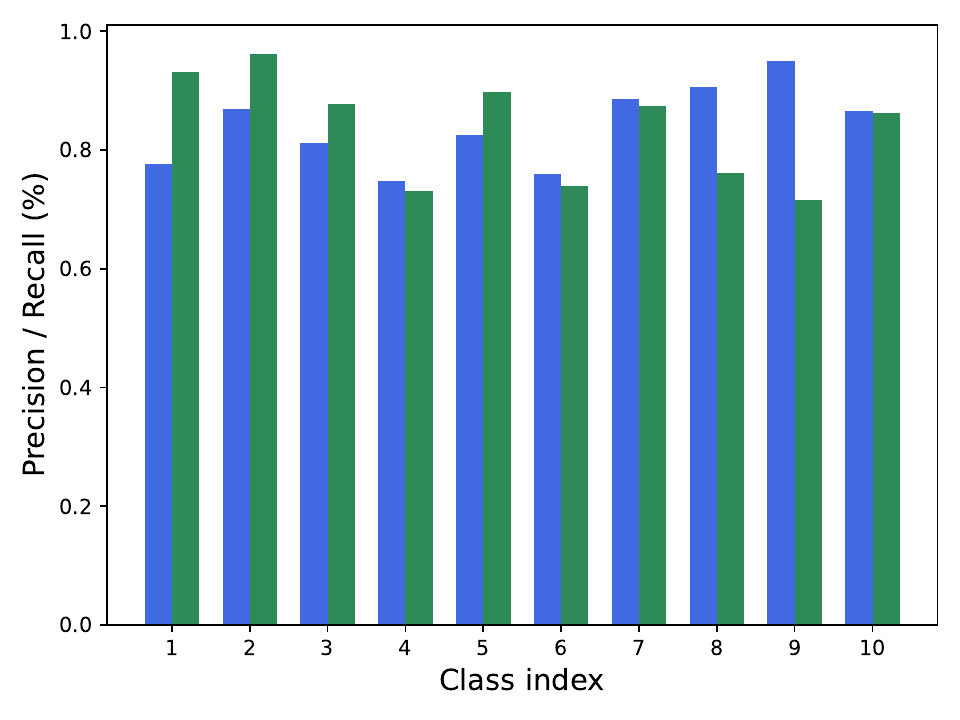}
    \caption{ACR w/BEM}
    \label{fig:image4}
  \end{subfigure}
  \caption{The precision and recall of the test set on CIFAR10-LT under $\gamma_l=\gamma_u$ setup.}
  \label{fig:prerecall}
\end{figure}
 
\noindent\textbf{Visualization of train curves and test accuracy class distribution.}
We further assess the effect of BEM on FixMatch and ACR by plotting training curves and class-wise test accuracy. As shown in  Fig.~\ref{fig:dis_acc}(a), the low entropy ratio increases, suggesting a large fraction of unlabeled data is used in the mixing as the training state becomes stable. As shown in Fig.~\ref{fig:dis_acc}(b), our method greatly improves the tail class performance of FixMatch and ACR.

\begin{figure}[!t]
\centering
\subfloat[Train curves]{\includegraphics[scale=0.2]{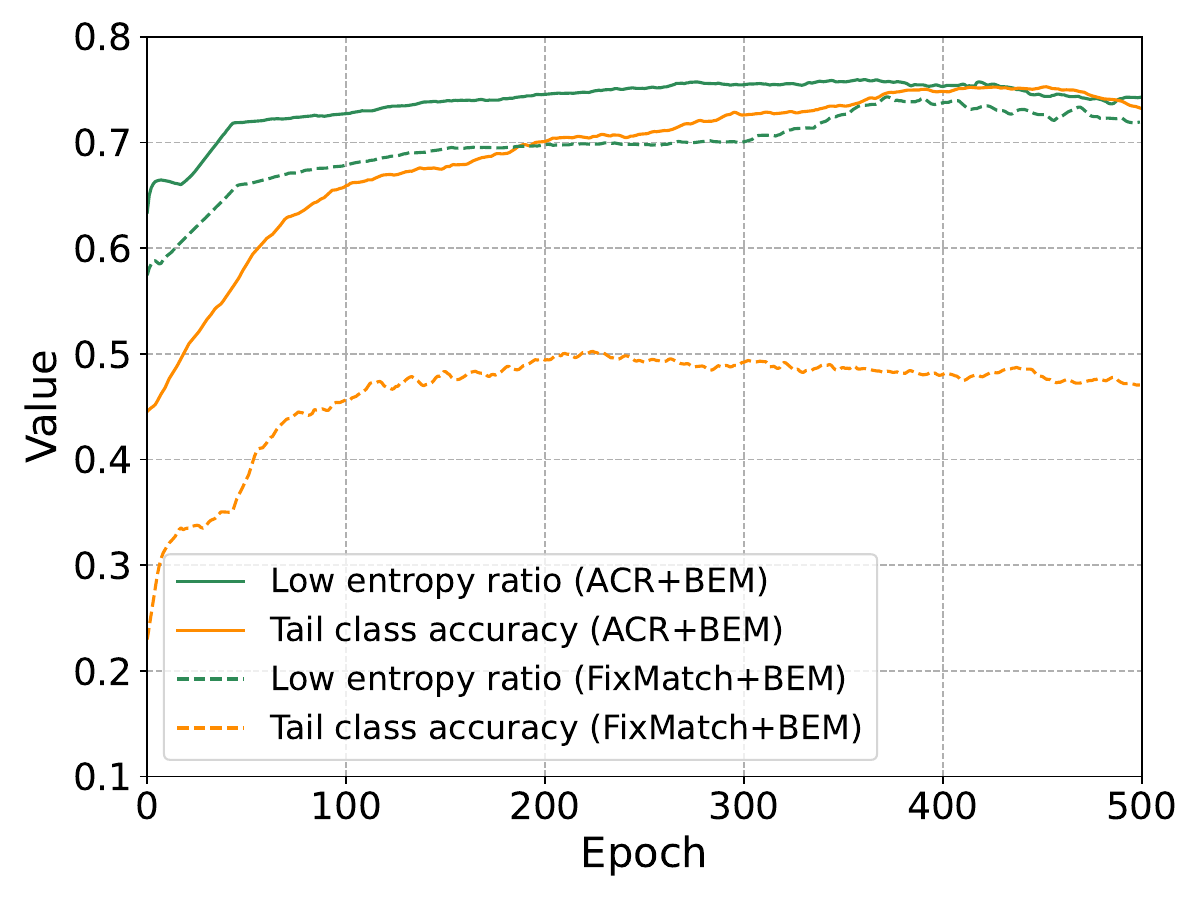}}
\subfloat[Class-wise test accuracy]{\includegraphics[scale=0.2]{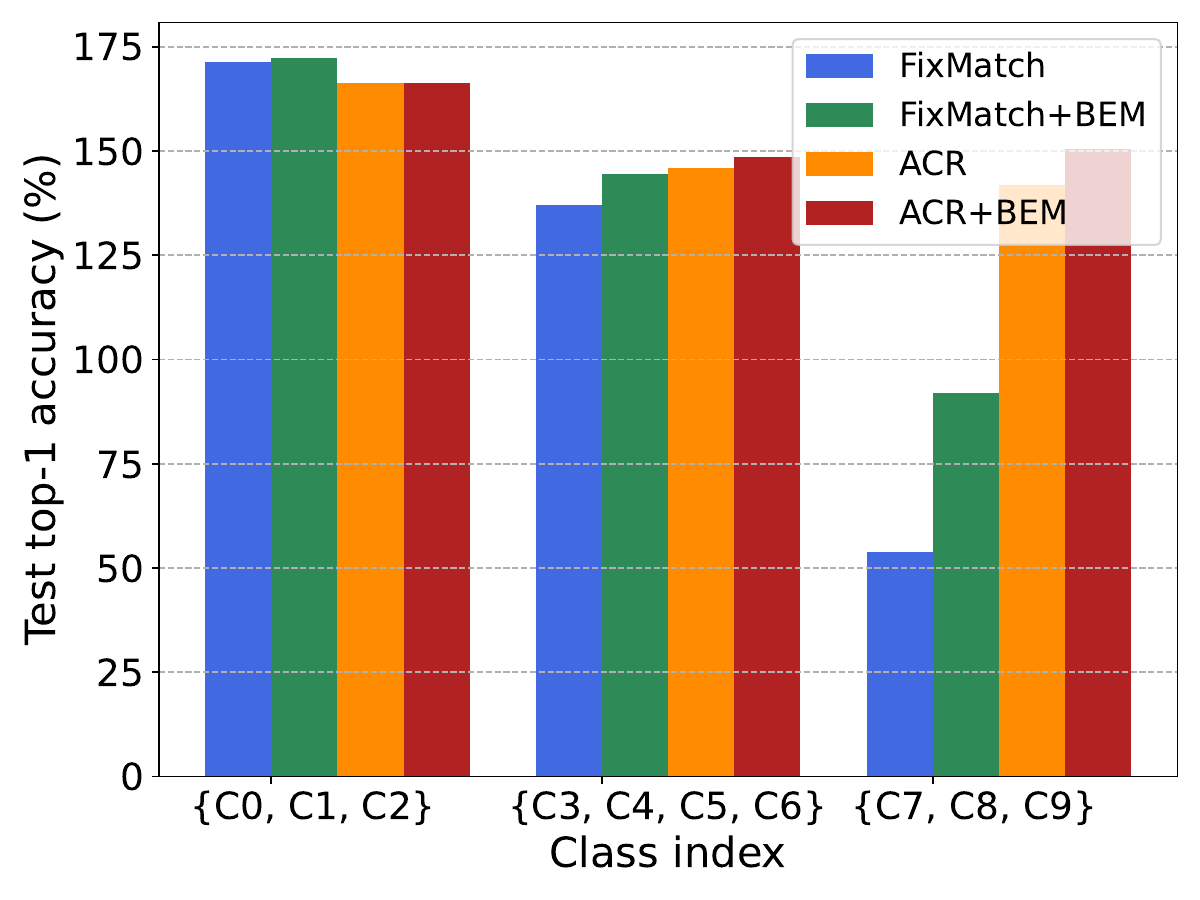}}\\
\caption{\textbf{(a):} Train curves for tail low entropy ratio and tail class accuracy. \textbf{(b):} Class distribution of test accuracy over different methods. C0 and C9 are the head and tail classes, respectively.  }
\label{fig:dis_acc}
\end{figure}



\noindent\textbf{Visualization of the class distribution of sampling rate and accuracy under different $\alpha$.}
We present the ablation study on $\alpha$ in Tab.~\ref{tab:alpha}. In addition, we further visualize the class distribution of sampling rate and accuracy under various $\alpha$. 
Fig.~\ref{fig:alpha_sampling_acc} (a) shows that as $\alpha$ increases, the sampling rate of tail classes improves. When $\alpha$ is small, the sampling function pays attention not only to tail classes but also to middle classes with high uncertainty. In Fig.~\ref{fig:alpha_sampling_acc} (b), we can see that when $\alpha=0.5$, both the tail class and the middle class with high uncertainty have relatively high accuracy, indicating it achieves the balance of data quantity and uncertainty.

\begin{figure}[!t]
\centering
\subfloat[Class-wise sampling rate ]{\includegraphics[scale=0.2]{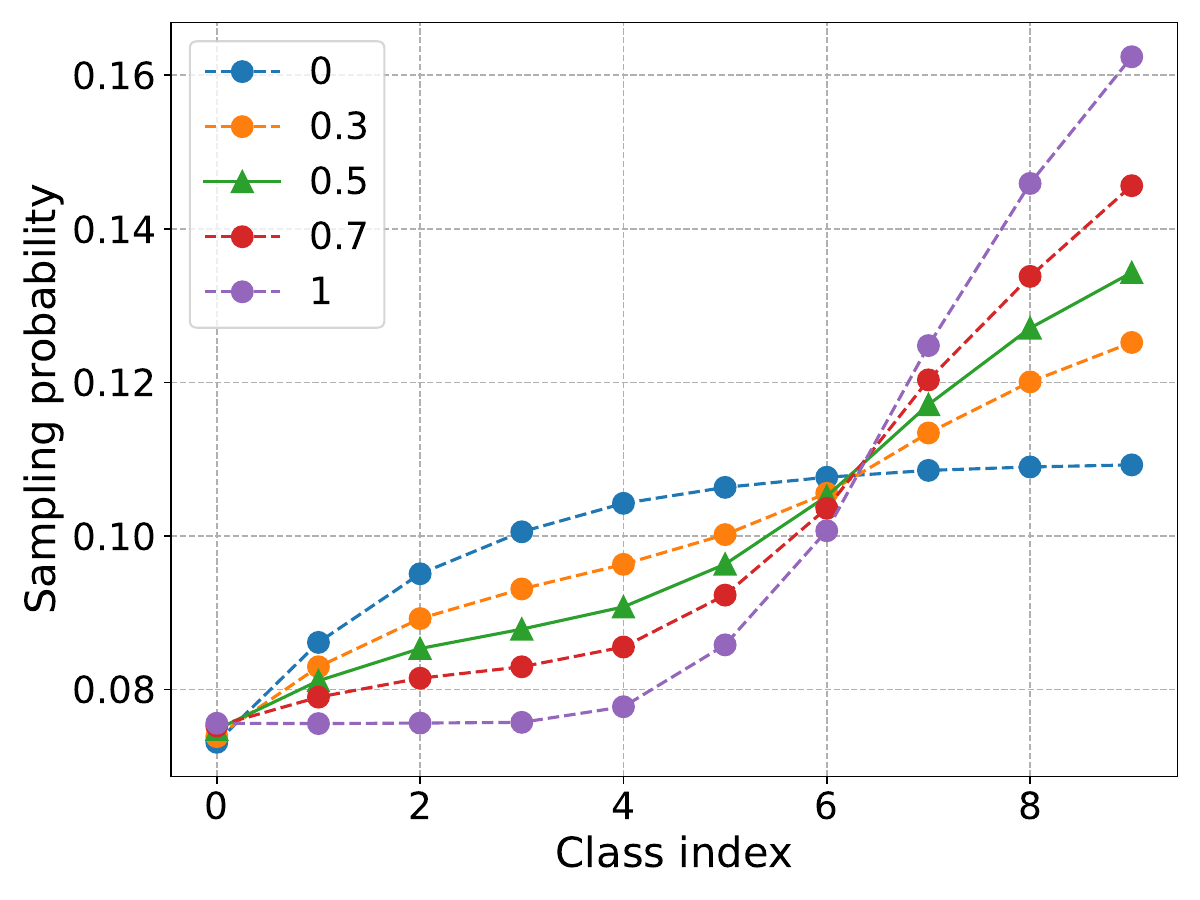}}
\subfloat[Class-wise test accuracy]{\includegraphics[scale=0.2]{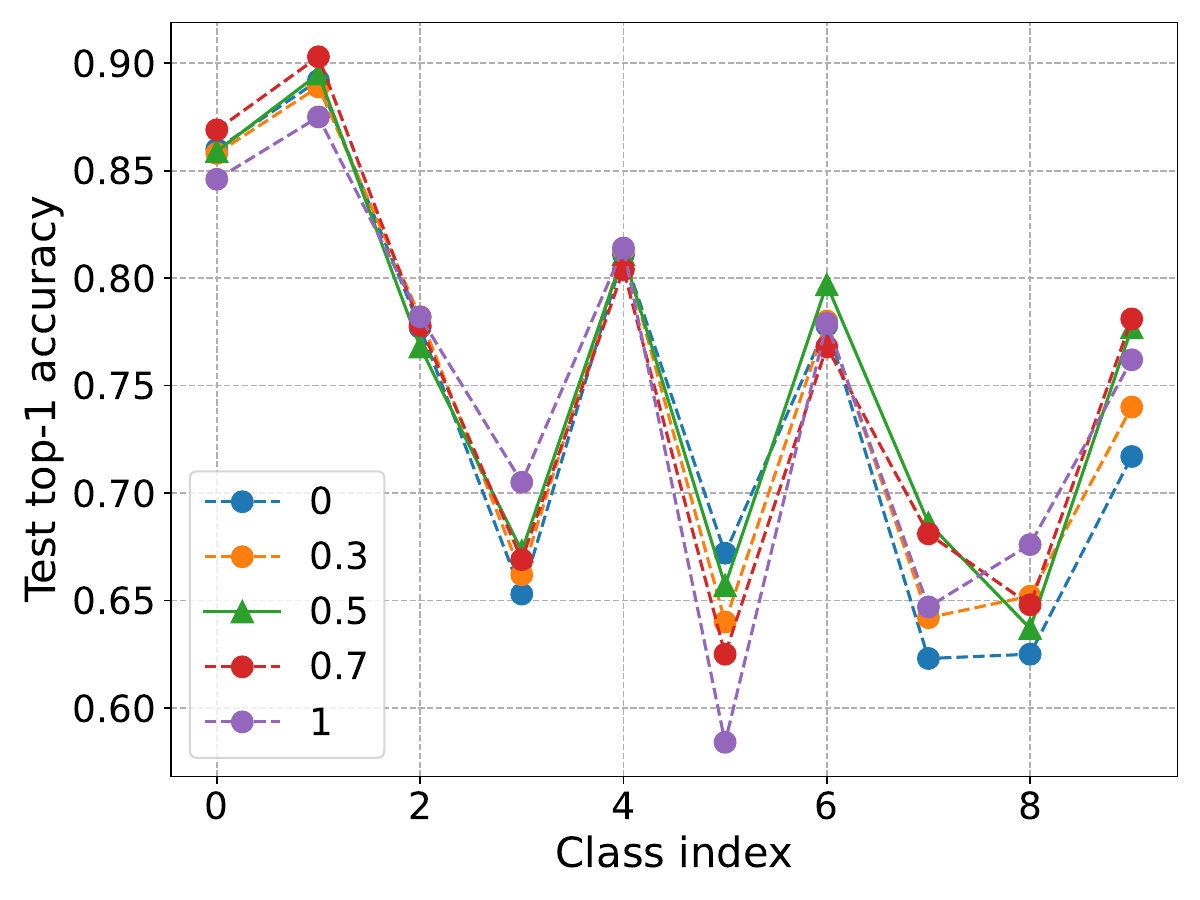}}\\
\caption{Class distribution of sampling rate and test accuracy under various $\alpha$ on CIFAR10-LT ($\gamma_l=\gamma_u=100$) using FixMatch.}
\label{fig:alpha_sampling_acc}
\vspace{-10pt}
\end{figure}

\begin{figure}[!t]
    \centering
    \includegraphics[width=0.99\linewidth]{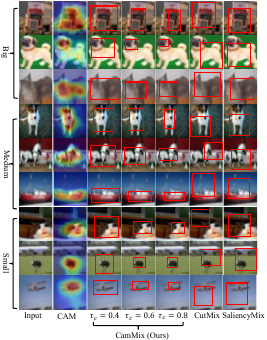}
    \caption{The visualization of data mixing process for CutMix, SaliencyMix, and CamMix on CIFAR10-LT. The red box indicates the image area selected by data mixing.}
    \label{fig:vis mix cifar10}
    \vspace{-10pt}
\end{figure}

\noindent\textbf{More visualization of data mixing.}
 We provide the intermediate images of the data mixing on STL10 in Fig.~\ref{fig:vis mix1}. To further illustrate the effectiveness of our CamMix, we also present additional visualization results on CIFAR10 in Fig.~\ref{fig:vis mix cifar10}. We select three images for each target size. Based on the results from the two datasets, we can draw the following conclusions: 1) CutMix has a high degree of randomness and often selects the context region. 2) The localization ability of SaliencyMix needs to be optimized. The selection region is not precise and tends to choose numerous redundant areas. 3) CamMix greatly improves the localization ability due to the accuracy of CAM and excludes irrelevant redundant areas as $\tau_c$ value decreases.

\noindent\textbf{More visualization of t-SNE}
As displayed in Fig.~\ref{fig:tsne}, we show the t-SNE of learning representations from the test data on CIFAR10-LT. We further conduct experiments on STL10-LT to visualize the learning representations when $\gamma_l \neq \gamma_u$. Results in Fig.~\ref{fig:tsne_stl10} show that our method generates clearer classification boundaries for representations when $\gamma_l \neq \gamma_u$.  
Specially, the classification ability of FixMatch is relatively poor, with most clusters gathered together. Our method greatly enhances its classification ability. 

\begin{figure}[!t]
\centering
\subfloat[FixMatch]{\includegraphics[scale=0.33]{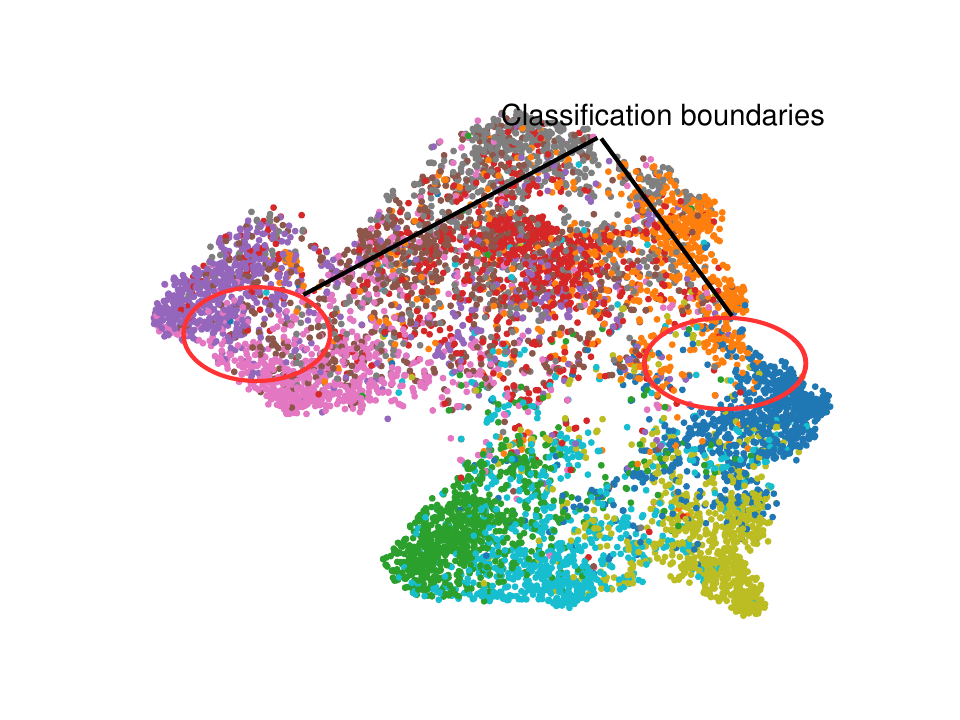}}
\subfloat[FixMatch w/BEM]{\includegraphics[scale=0.33]{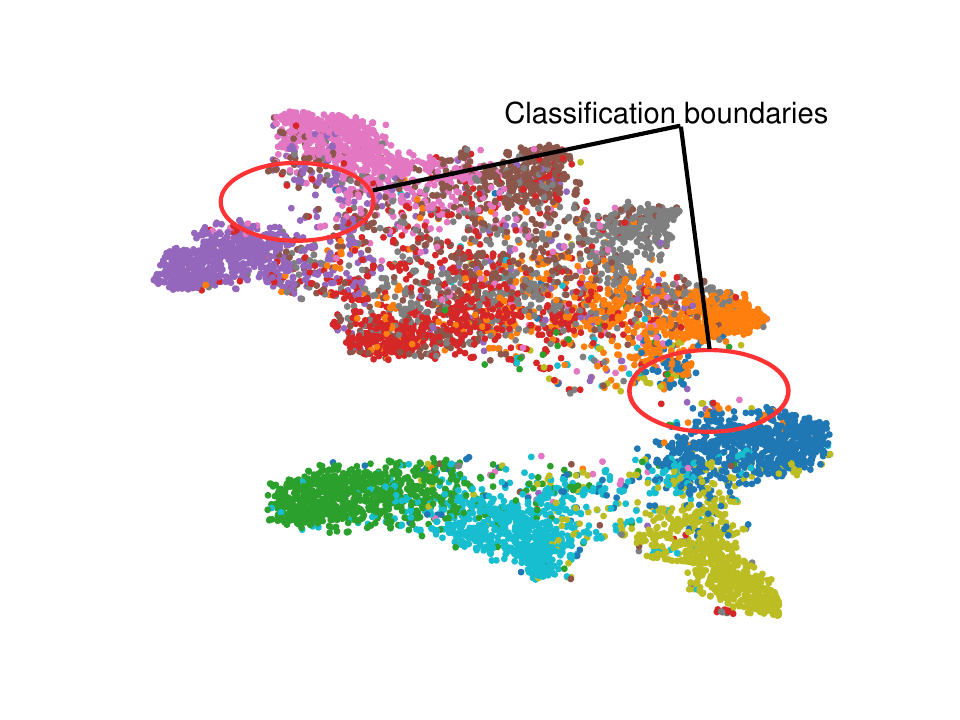}}\\
\subfloat[ACR]{\includegraphics[scale=0.33]{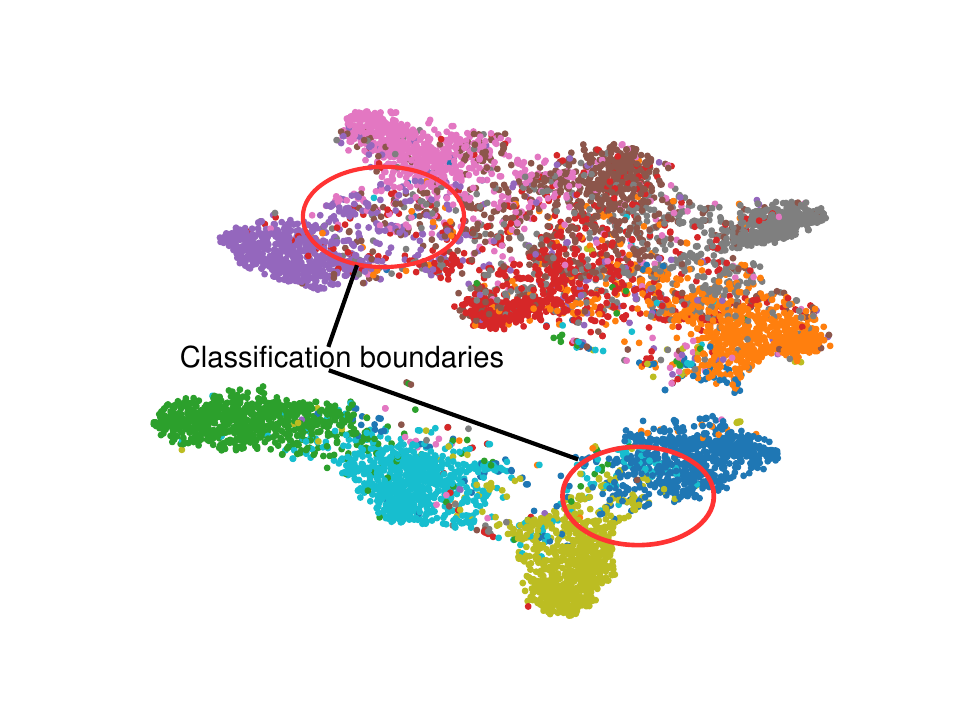}}
\subfloat[ACR w/BEM]{\includegraphics[scale=0.33]{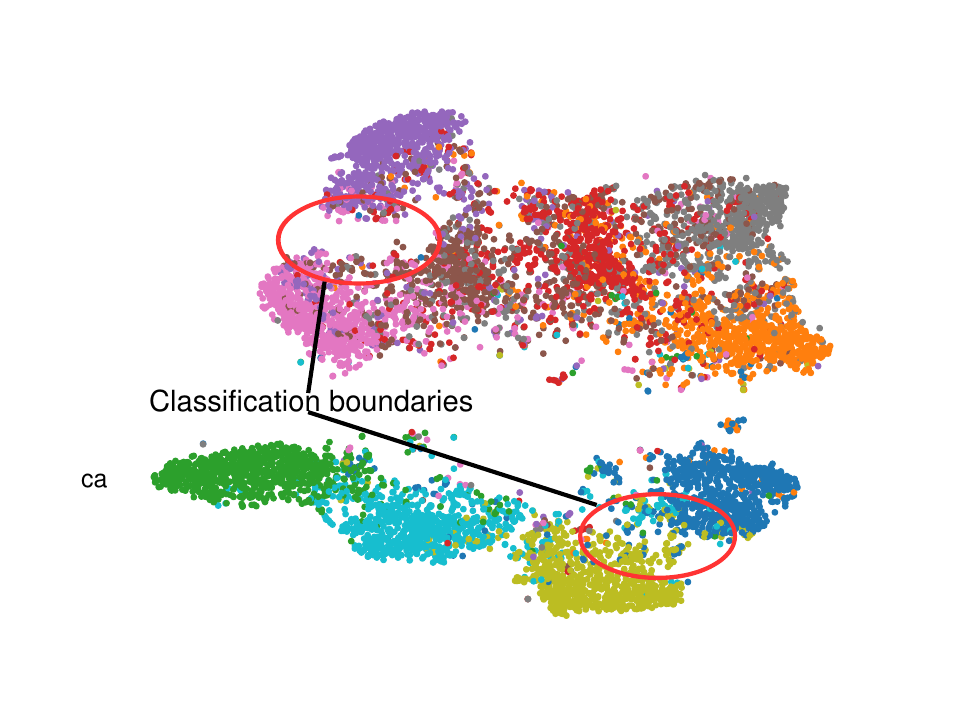}}\\
\caption{Comparison of t-SNE visualization with combinations of FixMatch and ACR on the test set of STL10-LT  when $\gamma_l\neq\gamma_u$. }
\label{fig:tsne_stl10}
\vspace{-10pt}
\end{figure}

\section{Limitation and Future Work} 
A potential limitation is that the proposed BEM is restricted by only exploring the data mixing for the LTSSL classification task, while ignoring its further application for other vision tasks, such as object detection~\cite{carion2020end,girshick2015fast,zhu2020deformable}, semantic segmentation~\cite{zheng2021rethinking,strudel2021segmenter,he2017mask,shi2022transformer} and others~\cite{li2023pose, zheng2023actionprompt, li2021hierarchical}. It is worth noting that the application of semi-supervised learning for long-tailed objection detection~\cite{zang2023semi,liu2020unbiased} and semantic segmentation~\cite{he2021re,hu2021semi} is not trivial but much harder than the pure classification task, as it requires further predict object location or semantic mask. In the future, we will extend our BEM to more complex vision tasks to further demonstrate its effectiveness and adaptability.

 \fi

\end{document}